\documentclass[10pt, a4paper, twocolumn, showabstract]{naverlabseurope}

\usepackage{lipsum}
\usepackage{multicol}
\usepackage{multirow}
\usepackage{graphicx}
\usepackage{caption}
\usepackage{tikz,tkz-kiviat,pgfplots}

\newlength{\smallimage}
        \definecolor{rel}{rgb}{.1,.6,.2}
        \definecolor{nrl}{rgb}{1,1,1}
        \definecolor{qim}{rgb}{1,1,1}

\def\ie{\emph{i.e.}}
\def\eg{\emph{e.g.}}

\definecolor{lightgray}{gray}{0.93}

\def\be{\begin{equation}}
\def\ee{\end{equation}}
\def\bea{\begin{eqnarray}}
\def\eea{\end{eqnarray}}
\def\ben{\begin{eqnarray*}}
\def\een{\end{eqnarray*}}

\def\bi{\begin{itemize}}
\def\ei{\end{itemize}}

\newcommand{\btab}[1]{\begin{tabular}{#1}}
\newcommand{\etab}{\end{tabular}}
\newcommand{\ba}[1]{\begin{array}{#1}}
\newcommand{\ea}{\end{array}}
\newcommand{\ul}[1]{\underline{#1}}

\def\Loss{\mathcal L}

\def\<{\langle}
\def\>{\rangle}

\newcommand{\bfB}{{\bf B}}
\newcommand{\bfC}{{\bf C}}
\newcommand{\bfD}{{\bf D}}
\newcommand{\bfE}{{\bf E}}

\newcommand{\bfI}{{\bf I}}

\newcommand{\bfM}{{\bf M}}

\newcommand{\bfX}{{\bf X}}

\newcommand{\bbfD}{{\bf \bar{D}}}

\newcommand{\calN}{{\mathcal N}}
\newcommand{\calP}{{\mathcal P}}

\newcommand{\R}{\mathbb{R}}

\newcommand{\myparagraph}[1]{\vspace{0.1cm}\noindent\textbf{#1.}}



\newcommand{\ith}{$i^\textrm{th}$\,}

\definecolor{DarkGreen}{rgb}{0.5, 0.9, 0.5}

\newcommand{\red}[1]{{\color{red}{#1}}}

\newcommand{\mypar}[1]{\noindent \textit{#1}\ }

\newcommand{\I}[1]{I_{#1}} 
\newcommand{\z}{z}           
\newcommand{\zgt}{\hat{z}}   

\newcommand{\dbl}{\textsc{Dec}} 
\newcommand{\cat}{\textsc{Cat}_\text{N}} 
 
\newcommand{\lin}{\textsc{Lin}} 
\newcommand{\head}{\textsc{Head}^{\text{3D}}}
\newcommand{\awa}{\textsc{Inj}^{\text{3D}}}

\newcommand{\lreg}{\ell_{\text{regr}}} 

\newcommand{\duster}[0]{DUSt3R}
\newcommand{\master}[0]{MASt3R}
\newcommand{\mastersfm}[0]{MASt3R-SfM}
\newcommand{\munster}[0]{MUSt3R}
\newcommand{\spanner}[0]{Spann3R}

\newcommand{\seq}{{\,{=}\,}}
\def\vs{\emph{vs.\,}}

\usepackage{pythonhighlight}

\title{\munster{}: Multi-view Network for Stereo 3D Reconstruction}

\titlerunning{\munster{}: Multi-view Network for Stereo 3D Reconstruction}

\correspondingauthor{[yohann.cabon,vincent.leroy]@naverlabs.com}

\authors{Yohann Cabon \,\,\,\,\,\,\,\,\,\,\, \authsep Lucas Stoffl \,\,\,\,\,\,\,\,\,\,\, \authsep Leonid Antsfeld \,\,\,\,\,\,\,\,\,\,\, \authsep Gabriela Csurka \,\,\,\,\,\,\,\,\,\,\, \authsep Boris Chidlovskii \,\,\,\,\,\,\,\,\,\,\, \authsep Jerome Revaud \,\,\,\,\,\,\,\,\, \authsep Vincent Leroy}
\affiliations{NAVER LABS Europe}
\contributions{} 
\website{https://europe.naverlabs.com}
\websiteref{\href{https://europe.naverlabs.com}}

\begin{abstract}
\duster{} introduced a novel paradigm in geometric computer vision by proposing a model that can provide dense and unconstrained Stereo 3D Reconstruction of arbitrary image collections with no prior information about camera calibration nor viewpoint poses. 
Under the hood, however, \duster{} processes image pairs, regressing local 3D reconstructions that need to be aligned in a global coordinate system. The number of pairs, growing quadratically, is an inherent limitation that becomes especially concerning for robust and fast optimization in the case of large image collections. 
In this paper, we propose an extension of \duster{} from pairs to multiple views, that addresses all aforementioned concerns. 
Indeed, we propose a Multi-view Network for Stereo 3D Reconstruction, or \munster{}, that modifies the \duster{} architecture by making it symmetric and extending it to directly predict 3D structure for all views in a common coordinate frame. Second, we entail the model with a multi-layer memory mechanism which allows to reduce the computational complexity and to scale the reconstruction to large collections, inferring thousands of 3D pointmaps at high frame-rates with limited added complexity. The framework is designed to perform 3D reconstruction both offline and online, and hence can be seamlessly applied to SfM and visual SLAM scenarios showing state-of-the-art performance on various 3D downstream tasks, including uncalibrated Visual Odometry, relative camera pose, scale and focal estimation, 3D reconstruction and multi-view depth estimation. Code will be available at \url{https://github.com/naver/must3r}.
\end{abstract}

\begin{document}
\maketitle

\section{Introduction}
\label{sec:intro}

Recently, \duster{} \cite{duster} introduced a novel paradigm in geometric computer vision.
In a nutshell, it can provide dense and unconstrained Stereo 3D Reconstruction of arbitrary image collections, \ie without any prior information about camera calibration nor viewpoint poses.
By casting the pairwise reconstruction problem as a regression of pairs of \emph{pointmaps}, where a pointmap is defined as a dense mapping between pixels and 3D points, it effectively relaxes the hard constraints of usual projective camera models. 
The pointmap representation, now used in subsequent works~\cite{master,mastersfm}, encompasses both 3D geometry and  the camera parameters, and allows to unify and jointly solve various 3D vision tasks such as  depth, camera pose and focal length estimation,  dense 3D reconstruction and pixel correspondences.
Trained from millions of image pairs with ground-truth annotations for depth and camera parameters, \duster{} shows unprecedented performance and generalization across various real-world scenarios with different camera sensors in zero-shot settings. 

\begin{figure}
    \centering
    \includegraphics[width=0.95\linewidth, trim=0 0 0 0, clip]{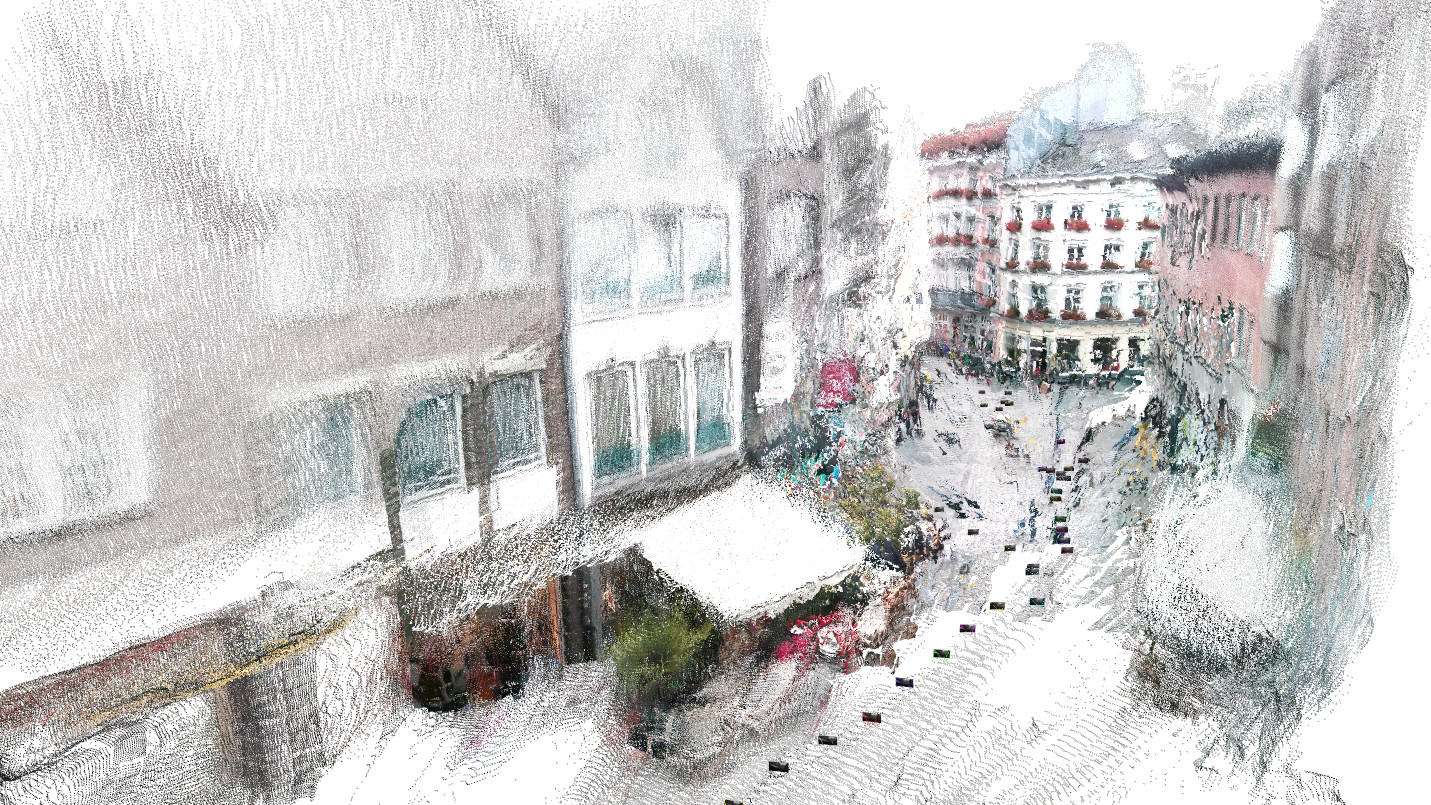}
    \hfill
    \includegraphics[width=0.95\linewidth, trim=0 0 0 0, clip]{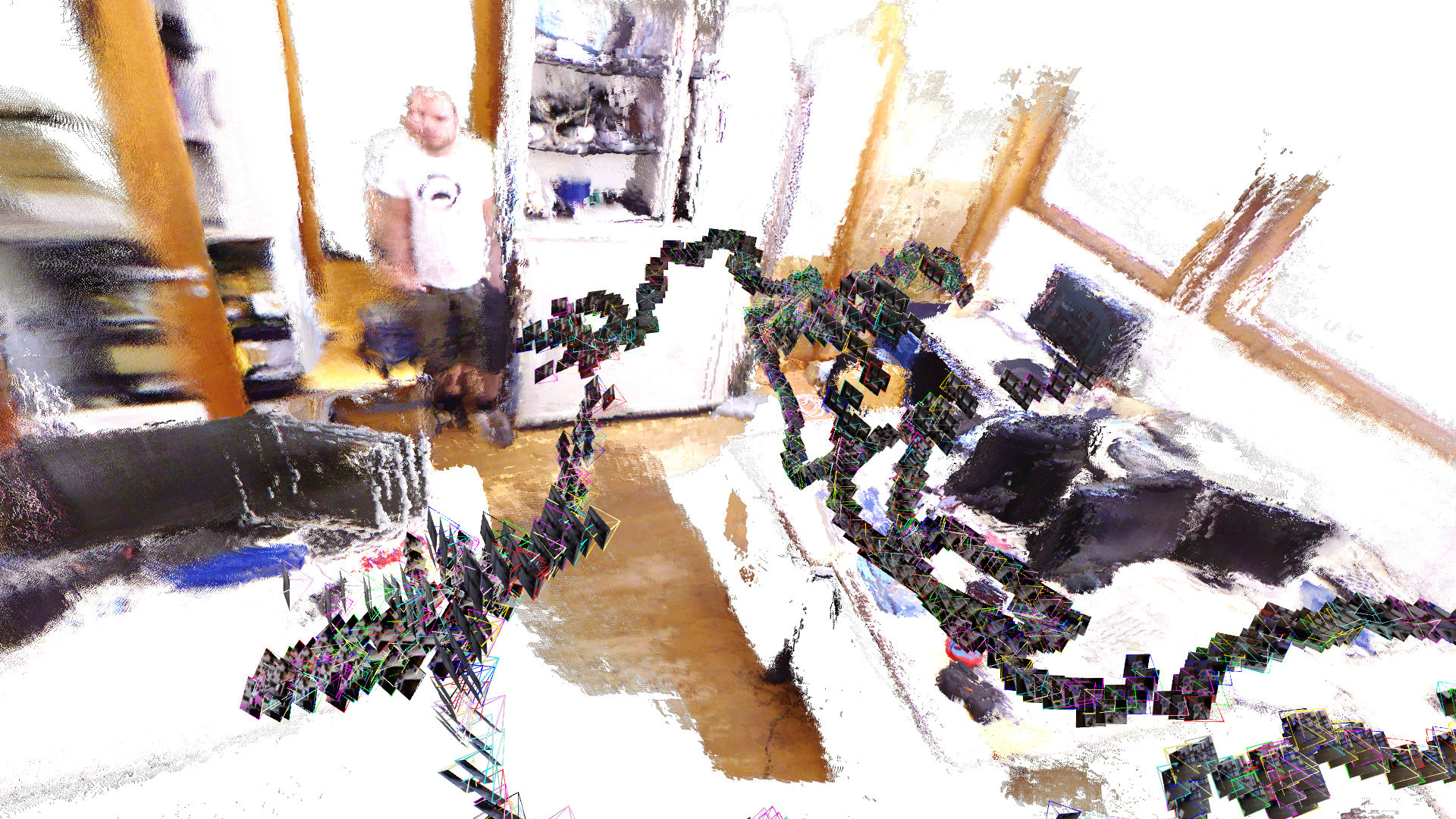}
    \caption{
        Qualitative example of \munster{} reconstructions of Aachen Day-Night \cite{aachen} nexus4 sequence 5 (offline, \textit{top}) and TUM-RGBD \cite{Sturm2012ASystems} Freiburg1-room sequence (online, \textit{bottom}). More qualitative examples can be found in Sec.~\ref{supsec:quali}. 
    }
    \label{fig:qualitative}
\end{figure}

 \begin{figure*}[ttt]
    \centering    
    \includegraphics[width=0.7\linewidth, trim=50 184 120 68, clip]{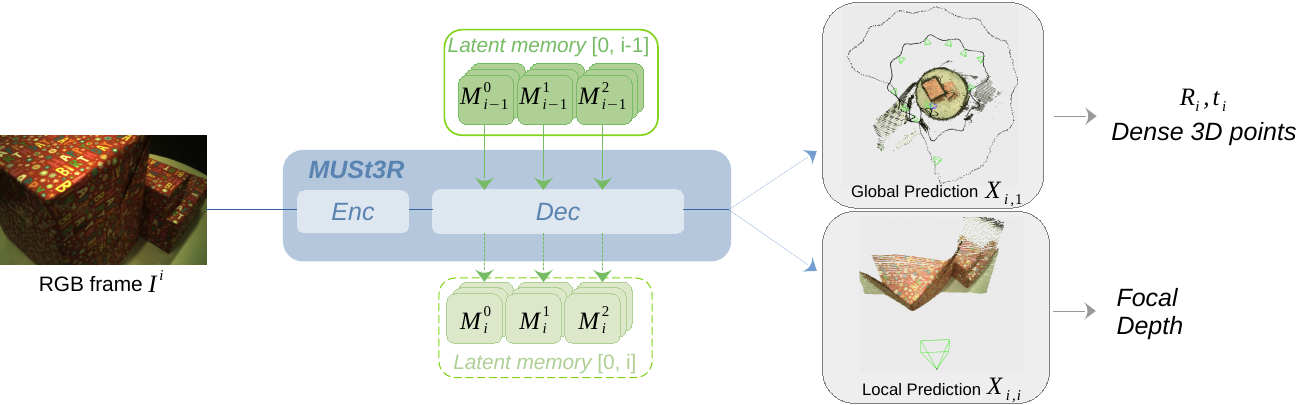}
    \vrule  
     \includegraphics[width=0.25\linewidth, trim=-100 0 0 0, clip]{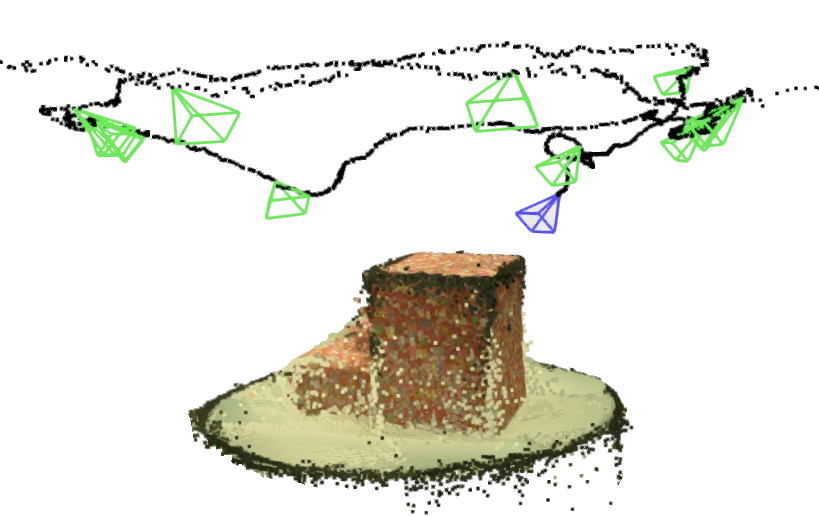}
    \caption{
        \textit{(Left)} \textbf{Overview of our uncalibrated reconstruction framework:} an input RGB, \munster{} architecture   and the memory state. The network predicts both local $\bfX_{i,i}$ and global $\bfX_{i,1}$ pointmaps, from which camera focal, depth map, pose and dense 3D can efficiently be recovered, as seen in the global reconstruction. The memory is optionally updated according to simple heuristics depending on the scenario. \textit{(Right)} \textbf{Qualitative example of uncalibrated Visual Odometry} on the ETH3D ``boxes'' sequence in the online setting.
    }
    \label{fig:overview}
\end{figure*}

Such architecture works seamlessly in monocular and binocular cases, yet when feeding many images, the pairwise nature of the approach becomes a drawback rather than a strength. 
Since the predicted pointmaps are expressed in a local coordinate system defined by the first image of each pair, all predictions live in different coordinate systems. 
This design hence requires a global post-processing step to align all predictions into one global coordinate frame, which quickly becomes intractable for large collections when done naively.
This raises the following questions: \textit{How to tackle the quadratic complexity of a pairwise approach? How can we robustly and quickly optimize such problem? What if we need real-time predictions? }
While these legitimate concerns are partially addressed in Mast3R-SfM~\cite{mastersfm}, we take here a different stance and design a new architecture that is scalable to large image collections of arbitrary scale, and that can infer the corresponding  pointmaps 
in the same coordinate system at high frame-rates. 
To achieve these, our \ul{Mu}lti-view network for \ul{St}ereo \ul{3}D \ul{R}econstruction (\munster{}),   extends  the \duster{} architecture through  several crucial modifications -- \ie making it symmetric and adding a working memory mechanism --   with limited added complexity. 

The model, beyond handling {\it offline} reconstruction of unordered image collections in a Structure-from-Motion (SfM) scenario, can also tackle the task of dense Visual Odometry (VO) and SLAM, which aims to predict {\it online} the camera pose and 3D structure of a video stream recorded by a moving camera. We present the first approach, to the best of our knowledge, that can seamlessly leverage the memory mechanism to cover both scenarios such that no architecture change is required and the same network can operate either task in an agnostic manner.    

Our contributions are threefold:
\begin{itemize}  
\item 
we revisit the \duster{} architecture by making it symmetric
    and enabling N-view predictions in metric space,
\item we entail it with a memory mechanism that allows to decrease the computational complexity
    for both offline and online reconstructions, and 
    \item we achieve state-of-the-art performance in both 
    unconstrained reconstruction scenarios in terms of estimating field-of-view, camera pose, 3D reconstruction and absolute scale without sacrificing any inference speed.   
\end{itemize}

\section{Related work}
\label{sec:related}

\myparagraph{Direct RGB-to-3D}
In contrast to traditional handcrafted approaches for Structure-from-Motion (SfM)~\cite{hartleymultiviewgeometry, jiang13,disco2013pami,cui2017hsfm, colmapsfm}, 
recent learning approaches aim at directly predicting 3D and cameras from one or several RGB images. These methods leverage neural networks to learn strong 3D priors from large datasets such as object-centric priors~\cite{shapefromsingleimg23,PavlloKHL21,PavlloSHML20,zero-1-to-3} or
arbitrary scenes~\cite{demon,deepv2d,deeptam,WangCVPR24VisualGeometryGroundedDeepSfM,smith24flowmap} via differentiable SfM trained end-to-end. \duster{}~\cite{duster} takes a different stance by casting the pairwise reconstruction task as a regression of a pair of pointmaps. This model was extended in~\cite{master} to improve its pairwise matching capabilities and by~\cite{zhang24monster} adapting it to dynamic scenes. In both cases however, the network only handles pairs, meaning that a Global Alignment (GA) needs to be introduced when more images are available. This comes with an additional complexity, a computational burden and technical challenges.

To avoid global alignment and increase inference speed,  \spanner{}~\cite{WangX24Spanner3DReconstructionWithSpatialMemory}, a concurrent work in the \duster{} framework proposes to use a spatial memory to keep track of previous observations. This allows to directly predict per-image pointmaps expressed in a global coordinate system, alleviating the need for GA.
\spanner{} retains the pairwise architecture of \duster{} processing pairs sequentially. Apart from the first and last one, all images go through the model twice successively. 
For each image, position-augmented features are obtained in the first pass, then enhanced by querying the spatial memory through an attention operation that involves an additional encoder. The second pass updates the memory and predicts the pointmap.
In this work, we divert from the pairwise paradigm and evolve \duster{} from pairwise to $N$-view by integrating the memory as an essential element of the architecture.
Interestingly, other works like ACE-0 \cite{brachmann2024acezero} also explored incremental pointmaps prediction in the form of scene coordinate reconstruction. 
The idea is to track image-to-scene correspondences and represents the scene implicitly as a neural network. Instead, our method represents scenes explicitly as pointmaps and deploys attention operations on memory for image-to-images correspondences.

\myparagraph{Calibrated Visual Odometry (VO)}
SLAM methods using only RGB cameras are less explored than their RGB-D counterparts, since the lack of geometric priors can cause scale and depth ambiguities. We refer the interested reader to a recent survey on SLAM in general~\cite{tosi2024nerfs} and we restrict ourselves to the VO framework in the following. The common denominator of all VO methods~\cite{teed21droid-slam,mur2017orb,ZhangX24GlORIESLAM} is a heavy reliance on handcrafted heuristics and projective camera geometry for joint camera pose and scene optimization, even when leveraging the most recent neural advancements~\cite{Rosinol2022NeRF-SLAM:Fields,chung2022orbeez,li2023dense,Zhu3DV24NICERSLAM}.
Most interesting to us are the works in the direction of prior-driven VO by including priors in the form of monocular depth~\cite{Zhu3DV24NICERSLAM,ZhouX24MoDSLAM}, monocular regularization~\cite{DexheimerCVPR23PCOMOCompactMappingOdometry} or learned optical flow~\cite{teed_dpvo_2023}. Likewise, our method leverages fully data-driven priors to operate online or offline indifferently. However, and in contrast to all existing VO methods, it can handle uncalibrated VO and can also seamlessly estimate a dense 3D scene, the camera parameters and scene scale at high inference speed.

\myparagraph{Uncalibrated Regime}
Dense VO without camera calibration is appealing for its scalability, versatility and ease of deployment. The problem has often been tackled in combination with other sensors, \eg multi-camera setups with IMU sensors~\cite{hengP2015selfcalibration}, or camera with LiDARs~\cite{rachman2023lidarcamera} and provably allows for efficient self-calibration.
RGB-only calibration~\cite{liao23cameracalibration_survey, fang22selfcalibration}, however, typically relies on CNNs trained with large amounts of synthetic data \cite{zhuang2019degeneracyselfcalibrationrevisiteddeep} and they are not often used in practice. An exciting option has been proposed with a fully uncalibrated solution to the navigation problem~\cite{koch2010navigation} but require a tedious preliminary exploration of a static environment. \munster{}, akin to \duster{} operates regardless of the camera parameters as long as it respects the camera models used for training. Interestingly, our dual pointmap prediction head enables efficient recovery of focal length to achieve high frame rates. 

\section{Method}
\label{sec:method}

We first briefly recall the main components of the \duster{} framework
in Sec. \ref{sec:duster}. Then, we describe the proposed \munster{} in Sec. \ref{sec:munst3r_base}
 and how it simplifies \duster{} to make it applicable to the N-view scenario. Our formulation enables the introduction of a memory mechanism, described in Sec. \ref{sec:causality}, that can be iteratively updated to handle an unlimited number of views. A major novelty of this approach lies in its ability to seamlessly tackle both offline reconstruction and sequential causal applications, such as dense Visual Odometry, at a high framerate. We describe how a single network can solve both scenarios in Sec. \ref{sec:memory}.

\subsection{\duster{}: a binocular architecture}
\label{sec:duster}
\duster~\cite{duster} is designed to jointly infer dense 3D reconstruction and camera parameters from pairs of images, by mapping a pair of images to 3D pointmaps that live in a  common coordinate system. A transformer-based network predicts a 3D reconstruction given two input images, in the form of two dense 3D pointmaps $\{\bfX_{i,1}\} \in \R^{H \times W \times 3}, i \in \{ 1, 2\} $ \ie a 2D-to-3D mapping between each pixel $p$ of the images $\{ I_i\}$ and the 3D point it observes $\bfX_{i,1}[p] \in \R^3$ expressed in the coordinate system of the first camera. 

Formally, given a pair of images $\{I_i\}$, they are first split into regular patches, or \emph{tokens}, that are encoded by a \emph{Siamese} ViT \cite{vit} encoder,  
yielding two latent representations $\bfE_i$.  
These are projected linearly to $\bfD^0_i=\lin(\bfE_i)$ which is 
the input to a set of $L$ intertwined layers of decoders blocks  $\{\dbl^l_1,\dbl^l_2\}_{l=1}^{L}$ similar to the \cite{croco} architecture. These blocks process the two images jointly, exchanging information via cross-attention at each layer to `{understand}' the spatial relationship between viewpoints and the global 3D geometry of the scene. Finally, two prediction heads $\{\head_i\}$ regress the final pointmaps $\bfX_{i,1}$ and their associated confidences $\bfC_i$ from the output of the last layers $\bfD^L_i$, and optionally $\bfE_i$, typically leveraged in combination with DPT prediction heads~\cite{dpt}:
\begin{align}
\bfX_{i,1},\bfC_i= \head_i(\bfE_i,\bfD^L_i)
\end{align}

\duster{} is trained in a fully-supervised manner using a simple pixel-wise regression loss
\begin{equation}
    \lreg(i,j) =\sum_{p \in I_i}\left\Vert \frac{1}{\z}\bfX_{i,j}[p]  - \frac{1}{\zgt}\widehat{\bfX}_{i,j}[p] \right\Vert, \label{eq:reg}
\end{equation}
where $j{=}1$ represent the reference view and $p$ is a pixel for which the ground-truth 3D point $\widehat{\bfX}_{i,j}[p] \in \R^3$ is defined. In the original formulation~\cite{duster}, normalizing factors $\z, \zgt$ are introduced to make the reconstruction scale-invariant.
They are defined as the mean distance of all valid 3D points to the origin. 
In this work we instead follow \master{}~\cite{master} and regress metric predictions when possible \ie we  set $\z := \zgt$ whenever ground-truth is metric. Following \duster~\cite{duster}, we also wrap this loss with a confidence aware loss $\Loss_{conf}$. 

\subsection{\munster{}: a Multi-view Architecture}
\label{sec:munst3r_base}
Our first contribution is to extend \duster{} to an arbitrary number $N$ of views. As detailed before, its binocular architecture features $2$ distinct decoders.
Naively extending to $N$ views would not scale, as it would practically require a set of $N$ distinct decoders. We propose instead to reformulate and simplify the previous framework by making the architecture symmetric with a single \emph{Siamese} decoder that shares weights between views. This architecture naturally scales to $N$ views while halving the number of trainable parameters in the decoder compared to that of \duster. 
Finally, we extend \duster{} to predict an additional pointmap that will be leveraged for efficient camera parameters estimation. We now detail each modification below.

\myparagraph{Simplifying the \duster{} architecture} 
First, we make the \duster{} architecture symmetric. 
Our hypothesis is that the duplicated decoders and heads are highly redundant. 
We thus replace them by a \emph{Siamese} decoder and a \emph{Siamese} head with shared weights, denoted as
$\dbl$ and $\head$ respectively, dropping the subscript notation. To identify the reference image $\I{1}$, which defines the common coordinate system, we add a learnable embedding $\bfB$ to $\bfD^0_2$ at the beginning of the shared decoder,
\begin{align}\bfD^0_2 = \lin(\bfE_2) + \bfB. \end{align} 
We also note that using Rotary Position Embedding (RoPE)~\cite{rope} in the cross attention is not necessary and can be safely removed. Please refer to Sec.~\ref{sec:archi} for detailed ablations on these changes.

\begin{figure}[ttt]
    \centering
    \includegraphics[width=0.56\linewidth,trim=50 120 580 20, clip]{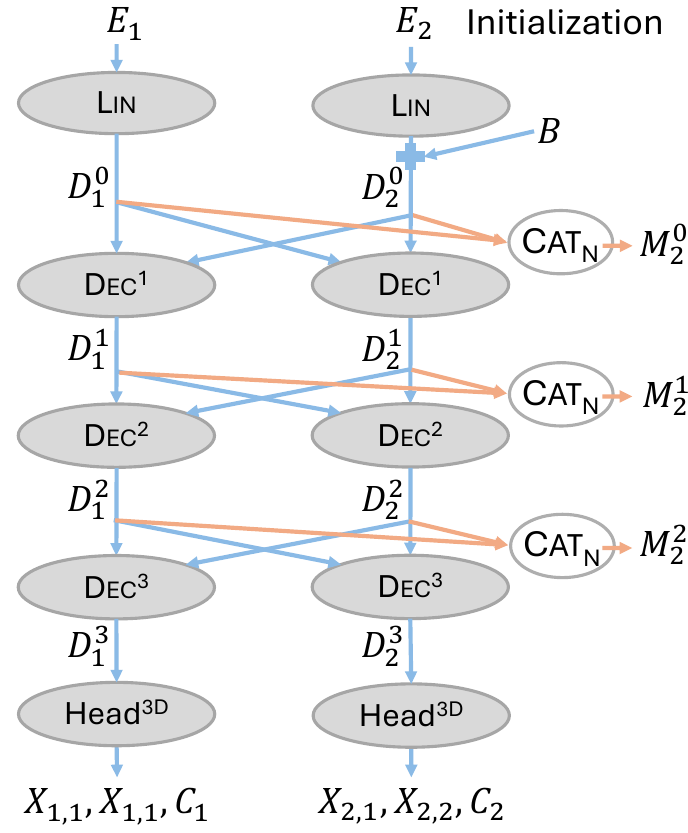}
    \vrule
    \includegraphics[width=0.41\linewidth,page=2,trim=150 120 570 20, clip]{figures/munst3r/munst3r_base_v4.pdf}
    \caption{
        Overview of the proposed architecture for a decoder of depth $L=3$, a Linear $\head$ and without the $\awa$ module.
        The left side shows initialization with two images. The right side shows how the memory is used and updated given a new image/frame. 
    }
    \label{fig:munst3r_arch}
\end{figure}

\begin{figure}[ttt]
    \centering
    \includegraphics[width=0.8\linewidth, trim=10 380 620 0, clip]{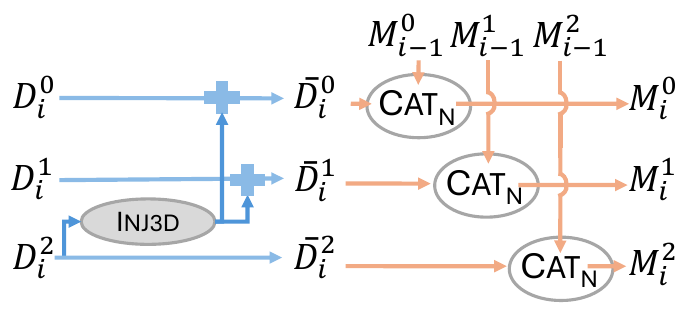}
    \caption{
        The 3D feedback module for a decoder of depth $L=3$.
    }
    \label{fig:munst3r_offset}
\end{figure} 

\myparagraph{Scaling-up to multi-view}
Our framework naturally extends to handle three or more images. This can be simply done by changing the behavior of the cross-attention in each decoder block $\dbl^l$. They are all residual and include self-attention (intra-view), followed by cross-attention (inter-view), and a final MLP. Therefore, we can trivially let the cross-attention operate between tokens of image $\I{i}$ and tokens of all other $j {\neq} i$ images. In more details, let $\cat$ denote the concatenation of image tokens in the sequence dimension and $\bfM^l_n = \cat(\bfD^l_1,\ldots,\bfD^l_{n})$
the concatenation of tokens from $n$ images at each layer $l$. Similarly, $\bfM^l_{n,-i} =\cat(\bfD^l_1,\ldots,\bfD^l_{i-1}, \bfD^l_{i+1},\ldots,\bfD^l_{n})$ denotes the concatenation of tokens for all but the \ith image. In this notation, our model applies, at each layer $l$, cross-attention between tokens of image $\I{i}$ and tokens of all other images:
\begin{align}
\bfD^{l}_i = \dbl^{l}(\bfD^{l-1}_i, \bfM^{l-1}_{n,-i} ).
\end{align}

\myparagraph{Fast relative pose regression}
In \duster{}, $\bfX_{1,1}$ is used to estimate the intrinsics of $\I{1}$, and a second forward with the symmetric pair $(\I{2}, \I{1})$ allows to predict $\bfX_{2,2}$ in order to estimate the intrinsics of $\I{2}$. We want to build a multi-view model that preserves this ability with a low computational cost. For this purpose we change the prediction head to output an additional $\bfX_{i,i}$ pointmap:
\begin{align}
(\bfX_{i,1},\bfX_{i,i},\bfC_i) = \head(\bfD^L_{i}), i \in \{1...n\}.
\end{align}
With such a change, we can easily recover the relative pose between $\I{1}$ and $\I{i}$ by estimating the transformation between $\bfX_{i,i}$ and $\bfX_{i,1}$ via Procrustes analysis, which is simpler and faster than PnP, as demonstrated empirically. Interestingly, we do so regardless of the focal length, in contrast to a more traditional PnP~\cite{hartleymultiviewgeometry} approach.

\subsection{Introducing Causality in \munster{}}

\label{sec:causality}
Our second contribution is to make \munster{} iterative. Based on the architecture described in~Sec. \ref{sec:munst3r_base}, we i) endow the model with an iteratively updated memory which allows to efficiently process any number of images, offline or online, ii) inject 3D feedback to earlier layers through an extra MLP. The overall decoder architecture is shown in Fig.~\ref{fig:munst3r_arch}, the injection schema is shown in Fig.~\ref{fig:munst3r_offset}.

\myparagraph{Iterative memory update}
In Sec. \ref{sec:munst3r_base}, we described how to extend 
\duster{} to process multiple images. Yet in practice $N$ may be very large, making cross-attention on very large token sequences computationally intractable. Furthermore, in some scenarios the images might arrive sequentially, for instance in visual odometry. In order to handle a large number of images we propose to leverage our model in an iterative manner, with the usage of a memory, reminiscent of \spanner{}~\cite{WangX24Spanner3DReconstructionWithSpatialMemory}. 
Contrary to \spanner{}, however, this memory simply contains in our case the previously computed $\bfM^{l}_n$ of every layer. As shown in Fig.~\ref{fig:munst3r_arch}, when a new image $I_{n+1}$ comes, it cross-attends with these saved tokens, \ie for each layer we have:
\begin{align}
\bfD^{l}_{n+1}   = \dbl^{l}(\bfD^{l-1}_{n+1}, \bfM^{l-1}_n) .
\label{eq:iterative}
\end{align}
Features $\bfD^{l}_{n+1}$ of the new image can simply be added to the memory by concatenating them to the current memory $\bfM^l_{n}$, thus expanding the memory to $\bfM^l_{n+1}$. Interestingly, we can draw a parallel with KV cache in causal transformer inference \cite{PopeMLSYS23EfficientlyScalingTransformerInference}. By caching the previously computed $\bfD^{l}_i$ at every layer, we make \munster{} causal: every new image attends to previously seen images, but these are not updated. 

With this architecture, it is possible to process an image without appending new tokens to the memory. We call this process \emph{rendering}. It can be used to break the causality of the model, \ie by re-computing pointmaps given tokens of future frames.
We typically perform rendering at the end of a video sequence, when all images are in the memory. 
We can process frames one by one (denoted as sequentially) or $n$ by $n$ ($n>1$), although sequential predictions usually perform better as shown in Sec.~\ref{sec:runtime}.

\myparagraph{Global 3D Feedback}
So far, the proposed method lacks any feedback mechanism between the memory tokens $\bfM^l_{n}$ of terminal layers towards those of earlier layers $\bfM^k_{n}$, $k<l$. 
In particular, $\bfM^{0}_{i}$ is just the concatenation of projected encoder features $\bfD^0_i$ and naturally lacks any knowledge of the other frames. 
One reasonable assumption is that the token representations at the terminal layer contain more global 3D information than those at earlier layers. 
We thus propose to simply augment all memory tokens with information from the last layer $l=L-1$ in order to propagate global 3D knowledge to every layer. This is feasible in the iterative framework described above since the last layers of the past frames already contain this information. Formally, let us denote the set of previous and new images by $\calP$ and $\calN$, respectively. To inject such information from the terminal layer into the earlier layers, we augment $\bfM^l_n$ with $\bfM^l_n = \cat(\bbfD^l_0,\ldots,\bbfD^l_n)$ 
where
\begin{align}
\bbfD^l_i = 
  \begin{cases} 
  \bfD^{l}_i+ \awa(\bfD^{L-1}_i),  & \forall l < L-1\,\text{and}\,i\in \calP \\
  \bfD^{l}_i,                      & l = L-1\, \text{or}\,i \in \calN,  \\
  \end{cases}\label{eq:awa2}
\end{align}
\noexpand
where $\awa$ consists of a Layer Norm followed by a two-layer MLP (see Fig. \ref{fig:munst3r_offset}).
We refer to empirical evidence in Sec.~\ref{sec:archi} that show the significant impact of this feedback mechanism.

\subsection{Memory Management}
\label{sec:memory}
As the memory grows linearly with the number of images, this can become an computational issue for large image collections. To mitigate this, we resort to a heuristic selection of memory tokens.
In fact, carefully selecting which images are added to the memory is crucial. 
We distinguish between two scenarios: online, where frames of a video stream come one by one, and offline, where we want to reconstruct an unordered collection of images. 
In all cases, we use the same network at test time without bells-and-whistles.

\begin{table*}
\begin{center}
\renewcommand\arraystretch{1.2}
\setlength{\tabcolsep}{1pt} 
\tiny
\hspace{-3mm}
\resizebox{1.6\columnwidth}{!}{
\begin{tabular}{cl|ccccccccccc|c|c}
\specialrule{1pt}{0.5pt}{0.5pt}
  & & \multicolumn{8}{c}{fr1} & \multicolumn{2}{|c|}{fr2} & \multicolumn{1}{c|}{fr3} & & Speed\\
  & & 360 & desk & desk2 & plant & room & rpy & teddy & xyz & \multicolumn{1}{|c}{xyz} & \multicolumn{1}{c|}{desk} & long & Avgerage & (FPS)\\
  \specialrule{1pt}{0.5pt}{0.5pt}
  \multirow{3}{*}{S} & ORB-SLAM3 \cite{campos_orbslam3_2021} & X & \textbf{2.0} & X & 11.8 & X & 5.6 & X & \textbf{1.0} & 0.5 & \textbf{1.3} & \textbf{1.7} & X &-\\
  & DSO \cite{engel_dso_2018} & X & 27.2 & 66.0 & 6.0 & 58.6 & X & X & 3.8 & 0.3 & 2.2 & 9.9 & X &-\\
  & DPVO \cite{teed_dpvo_2023} & 13.1 & 9.4 & 6.5 & \textbf{3.0} & 39.8 & 3.5 & 6.2 & 1.3 & 0.5 & 3.5 & 5.5 & 8.4 &-\\
\specialrule{1pt}{0.5pt}{0.5pt}
\multirow{7}{*}{D} & TANDEM \cite{koestler_tandem_2022} & X & 4.3 & 33.7 & X & X & 4.9 & 43.1 & 2.4 & 0.3 & 2.0 & 8.3 & X &-\\
  & MonoGS \cite{matsuki_gaussian_2024} & 14.2 & 6.3 & 74.0 & 9.3 & 64.9 & 3.4 & 35.6 & 1.6 & 4.5 & 133.1 & 3.3	& 31.8 &-\\
  & DeepFactors \cite{czarnowski_deepfactors_2020} & 17.9 & 15.9 & 20.2 & 31.9 & 38.3 & 3.8 & 56.0 & 5.9 & 8.4 & 26.3 & 49.0 & 24.9 &-\\
  & DepthCov \cite{dexheimer_learning_2023} & 12.8 & 5.6 & 4.8 & 26.1 & 25.7 & 5.2 & 47.5 & 5.6 & 1.2 & 15.9 & 68.8 & 19.9 &-\\
  & DROID-VO \cite{teed21droid-slam} & 15.7 & 5.2 & 11.1 & 6.0 & 33.4 & 3.2 & 19.1 & 5.6 & 10.7 & 7.9 & 7.3 & 11.4 &-\\
  & COMO-NC \cite{DexheimerCVPR23PCOMOCompactMappingOdometry}& 16.1 & 4.2 & 10.9 & 19.3 & 28.6 & 5.2 & 68.7 & 4.1 & 0.7 & 8.8 & 46.8 & 19.4 &-\\
  & COMO \cite{DexheimerCVPR23PCOMOCompactMappingOdometry} & 12.9 & 4.9 & 9.5 & 13.8 & 27.0 & 4.8 & 24.5 & 4.0 & 0.7 & 6.3 & 10.5 & 10.8 & -\\
  & GlORIE-VO* \cite{ZhangX24GlORIESLAM} & 13.1 & 4.0 & 8.6 & 4.1 & 32.7 & \textbf{2.9} & 14.5 & 1.2 & \textbf{0.2} & 16.1 & 4.8 & 9.3 &-\\
\hline
 & \spanner{} \cite{WangX24Spanner3DReconstructionWithSpatialMemory} & 20.7 & 16.1 & 28.3 & 57.4 & 84.8 & 6.1 & 92.4 & 2.1 & 4.4 & 20.7 & 193.9 & 47.9 & 4.8\\
U & \munster{}-C &  8.9 & 5.1 & 7.1 & 5.4 & 13.4 & 5.2 & 6.9 & 2.7 & 1.7 & 15.6 & 5.9 & 7.1 &\textbf{11.1}\\
 & \munster{} &  \textbf{7.8} & 4.0 & \textbf{4.6} & 4.0 & \textbf{9.9} & 4.3 & \textbf{4.2} & 1.3 & 1.2 & 15.3 & 4.3 & \textbf{5.5} & 8.4\\
\specialrule{1pt}{0.5pt}{0.5pt}
\end{tabular}
}
\normalsize
\caption{\textbf{VO: ATE RMSE [cm] on TUM RGB.} Sparse (S) versus dense (D) versus dense unconstrained (U) methods on TUM-RGB SLAM benchmark. (*) model re-run without Loop Closure and global bundle adjustment. 
}
\label{tab:tum_ate}
\end{center}
\end{table*}

\myparagraph{Online} 
In the video case, we leverage a running memory and 3D scene of current observations which are updated on-the-fly. The memory and the scene are initialized from the predictions of the first image. Then, we forward every incoming frame through \munster{}, attending to the current memory. 
This leads to a prediction of both dense visible geometry and camera parameters. 
We decide whether to keep the current prediction based on the spatial \emph{discovery rate} between the predicted pointmap $X_{i,1}$ and the current scene, keeping only a frame when it observes a significantly new part of the scene, or from a different enough viewpoint.\\
\indent To this aim, we store the scene as a set of KDTrees~\cite{BentleyCACM75MultidimensionalBinarySearchTrees}.
When building or querying the trees, each 3D point is associated to a tree by index based on the viewing direction of the observation. This is done by splitting the sphere of viewing directions into regular octants. We discretize the view direction of each pixel in spherical coordinates, to map it to the index of the relevant octant. 
Each pixel is thus mapped to a specific tree, then used to recover the nearest distance to the current scene. This distance is normalized by the depth at this pixel. 
The \emph{discovery rate} of a frame is simply the $p$\textit{-th} percentile of the normalized distances. We decide to add the frame to the memory and the 3D points and view directions to the current 3D scene if the \emph{discovery rate} is above a given threshold $\tau_d$, \ie the incoming frame observes enough new regions of the scene. 
\\
\indent An example of kept memory frames are shown in green in Fig.~\ref{fig:overview} (right).
We refer to Fig.~\ref{code:online} for the detailed algorithm. 
Note that this approach is purely causal since each view only sees the past frames, still we can break  the causality by \emph{rendering} again all images, as described in Sec. \ref{sec:causality}.

\myparagraph{Offline} 
Inspired by \mastersfm{} \cite{mastersfm}, we use the ASMK (Aggregated Selective Match Kernels) image retrieval method \cite{asmk} using the encoder features $\bfE_i$ of all images $\bfI_i$. 
Our key insight is to leverage the encoded images with minimal computational overhead. We follow their farthest point sampling~\cite{fps} method to select a fixed number of keyframes. 
The problem is then to find a good ordering of the images such as to observe the ones that maximize the overlap first, for more stability in the predictions. 
Therefore, we reorder them with the following strategy: we start with the keyframe which is the most connected to the others, then a greedy loop iteratively adds the other images by order of highest similarity to the current view set. 
These keyframes are sequentially passed through the network to build a latent representation of the whole scene. We then \emph{render} all the images from this memory. Note that it is 
possible to forward all images in an iterative manner like in ACE-0~
\cite{brachmann2024acezero}, but this would dramatically increase the number of decoder passes needed, thus adding computational burden.

\section{Training}
\label{sec:details}
\myparagraph{Pre-training \munster{} with pairs} 
Similar to \duster{}, we train \munster{} in multiple steps. First, we start by training the simplified architecture described in Sec.~\ref{sec:munst3r_base} for metric predictions. In our scenario, we aim to predict points that could be far apart in a large scene. 
For a better convergence and performance on distant points, we compute it in log space (see the ablation study of the loss choices in Sec.~\ref{sec:archi}) : 
\begin{align}
&f: x \xrightarrow{} \frac{x}{\left\Vert x \right\Vert}log(1+\left\Vert x \right\Vert),\\ 
&\bfX'_{i,j}[p] = f(\frac{1}{\z}\bfX_{i,j}[p]),\,\, \widehat{\bfX}'_{i,j}[p] = f(\frac{1}{\zgt}\widehat{\bfX}_{i,j}[p]),\\
&\lreg(i,j) = \sum_{p \in I_i}\left\Vert \bfX'_{i,j}[p] - \widehat{\bfX}'_{i,j}[p] \right\Vert.\label{eq:log_reg}
\end{align}
We start training  the model with a linear head initialized from CroCo v2 \cite{croco_v2} (decoder depth $L=12$) on  224 resolution images. Then, we finetune for 512 resolution 
(with varying aspect ratios). The model is trained  on a mixture of 14 datasets:
Habitat~\cite{Savva_2019_ICCV}, 
ARKitScenes~\cite{arkitscenes}, 
Blended MVS~\cite{blendedMVS}, 
MegaDepth~\cite{megadepth}, 
Static Scenes 3D~\cite{MIFDB16}, 
ScanNet++~\cite{scannet++}, 
CO3D-v2~\cite{co3d}, 
Map-free~\cite{mapfree}, 
WildRGB-D~\cite{XiaCVPR24RGBDObjectsInTheWild},
Virtual KITTI~\cite{CabonX20VirtualKITTI2},
Unreal4K~\cite{TosiCVPR21SMDNetsStereoMixtureDensityNetworks},
DL3DV ~\cite{LingCVPR24DL3DV-10KLargeScaleSceneDataset},
TartanAir~\cite{WangIROS20TartanAir} and
an internal dataset.

\begin{table*}[ttt]
\begin{center}
\renewcommand\arraystretch{1.2}
\setlength{\tabcolsep}{1pt} 
\tiny
\resizebox{1.6\columnwidth}{!}{
\begin{tabular}{l|cccccc|c}
\specialrule{1pt}{0.5pt}{0.5pt}
               & Test & Handheld & Robot & Structure & Dynamic & 3D Objects & Avg \\
\specialrule{1pt}{0.5pt}{0.5pt}
\spanner{}~\cite{WangX24Spanner3DReconstructionWithSpatialMemory} 
              &4.1(4.0)&60.1(28.3)&122.8(115.6)&32.7(6.8)& 88.3(76.6)&80.9(61.2) & 64.8(48.7) \\
\munster{}-C&3.9(\textbf{3.8})&  7.2(7.1)&  37.4(32.6)& 5.2(5.0)& 39.1(12.7)& 32.0(7.1) & 20.8(11.4) \\
\munster{}&\textbf{3.8}(\textbf{3.8})&\textbf{6.2(5.8)}&\textbf{20.6(14.6)}&\textbf{3.9(3.6)}&\textbf{29.1(10.8)}&\textbf{22.7(5.6)}&\textbf{14.4(7.4)} \\ 
\specialrule{1pt}{0.5pt}{0.5pt}
\end{tabular}}
\normalsize
\caption{\textbf{TUM RGBD~\cite{Sturm2012ASystems} Tracking Accuracy} ATE RMSE [cm]. 
Dense unconstrained methods on the whole TUM RGBD dataset split into six categories: mean (median) RMSE values are reported per category.}
\label{tab:tum_group}
\end{center}
\end{table*}

\myparagraph{Training \munster{}}  Then, we train \munster{} with 
multiple views, starting from the above trained symmetric \duster{} 
as initialization. In our experiments we use a total number of $N=10$ images per scene\footnote{For object centric datasets, the $N$ views are randomly selected, while for the others, we start from an overlapping image pair and add images one by one, ensuring that it overlaps with at least one already selected image.}. To be able to handle such sequences of images, we freeze the encoder and use xformers \cite{xFormers2022} to compute attention efficiently. We train this model on 12 datasets, that are mentioned above, but removing Virtual KITTI and Static Scenes 3D as they are unsuitable to our setup. During training, the memory is initialized from two images, then updated from individual images, as shown in Fig.~ \ref{fig:munst3r_arch}. We split the training loss in two steps: 1) we predict the pointmaps of a randomly chosen number $n, 2 \leq n\leq N$ of views, and use the latent embeddings to populate the memory, and 2) we \emph{render} all views, including the $n$ memory frames from this memory, meaning we obtain in the end $n+N$ predictions that correspond to the concatenation of the $n$ and $N$ views. The loss to minimize is thus :
\begin{equation}
    \Loss = \sum_{i \in 1}^{n+N}\lreg(i,1) + \lreg(i,i). \label{eq:loss1}
\end{equation}

To increase robustness and favor redundancy, we augment the training with a token dropout. The memory tokens from the first image $\I{1}$ are protected as it plays a particular role for the 3D points are represented in the coordinates of the first camera, similar to \duster{}. Token dropping is made for each incoming frame on the current memory and is consistent across layers, such that if a token is removed, it should not appear in any layer. 
We use a dropout probability of 0.05 (0.15) for 224 (512) resolutions respectively. 

\section{Experimental validation}
\label{sec:exp}

We demonstrate the usability and performance of the presented system in many unconstrained scenarios, namely uncalibrated Visual Odometry (VO), relative pose estimation, 3D reconstruction and multi-view depth estimation. In all scenarios we achieve state-of-the-art performance without access to the camera calibration, yet with a pipeline of a striking versatility and simplicity. Our evaluations are comparing \munster{} to the existing state of the art on many downstream tasks (Sec.~\ref{ssec:relpose}, \ref{ssec:3dr}, \ref{ssec:mvd}), but we put a particular focus on \duster{} and \spanner{}, which are the closest to our work. We ablate the components and design choices of the \munster{} architecture in the Sec.~\ref{sec:archi}.

\begin{table}[ttt]
\begin{center}
\renewcommand\arraystretch{1.2}
\setlength{\tabcolsep}{1pt} 
\tiny
\resizebox{\columnwidth}{!}{
\begin{tabular}{l|ccccccccccc|cc}
\specialrule{1pt}{0.5pt}{0.5pt}
       & \multicolumn{8}{c|}{freiburg1} & \multicolumn{2}{c|}{freiburg2} & \multicolumn{1}{c|}{freiburg3} &&\\
       & 360   & desk & desk2 & plant & room  & rpy & teddy & xyz  & \multicolumn{1}{|c}{xyz} & desk  & \multicolumn{1}{|c|}{long\_office} & Mean & Median \\
\specialrule{1pt}{0.5pt}{0.5pt}

\spanner & 12.13 & 12.77 & 12.16 & 12.68 & 12.12 & 11.64 & 12.81 & 12.36 & 11.79 & 12.74 & 9.49 & 12.06 & 12.16\\
\munster & \textbf{6.42}&\textbf{2.64}&\textbf{3.36}&\textbf{6.56}&\textbf{4.32}&\textbf{4.04}&\textbf{5.65}&\textbf{0.11} & \textbf{5.06}&\textbf{6.28}&\textbf{3.04}&\textbf{4.32}&\textbf{4.32} \\
\specialrule{1pt}{0.5pt}{0.5pt}
\end{tabular}
}
\normalsize
\caption{\textbf{Comparison of vertical FoV errors} in degrees for \spanner{} and \munster{} across TUM RGBD scenes.}
\label{tab:tum_fov}
\end{center}
\end{table}

\begin{table}[ttt]
\begin{center}
\renewcommand\arraystretch{1.2}
\setlength{\tabcolsep}{1pt} 
\tiny
\resizebox{\columnwidth}{!}{
\begin{tabular}{l|ccccccccccc|cc}
\specialrule{1pt}{0.5pt}{0.5pt}
       & \multicolumn{8}{c|}{freiburg1} & \multicolumn{2}{c|}{freiburg2} & \multicolumn{1}{c|}{freiburg3} & &\\
       & 360   & desk & desk2 & plant & room  & rpy & teddy & xyz  & \multicolumn{1}{|c}{xyz} & desk  & \multicolumn{1}{|c|}{long\_office} & Mean & Median\\
\specialrule{1pt}{0.5pt}{0.5pt}
\munster-C & 26.9 &3.3 & 9.8 & \textbf{1.0} & 7.9 & \textbf{85.9} & \textbf{14.1} & 3.7 & \textbf{1.2} & 2.4 & 5.5 & 14.7 & 5.5 \\
\munster & \textbf{21.1} & \textbf{2.1} & \textbf{9.3} & 2.9 & \textbf{7.7} & 86.3 & 15.2 & \textbf{0.2} & 1.7 & \textbf{1.7} &\textbf{4.6} & \textbf{13.9} & \textbf{4.6}\\
\specialrule{1pt}{0.5pt}{0.5pt}
\end{tabular}
}
\normalsize
\caption{\textbf{Scale estimation error} (in \% of the \emph{ground-truth} scale) across TUM-RGBD scenes.}
\label{tab:tum_scale}
\vspace{-3mm}
\end{center}
\end{table}

\begin{table}[ttt]
\begin{center}
\renewcommand\arraystretch{1.2}
\setlength{\tabcolsep}{1pt} 
\tiny
\resizebox{\columnwidth}{!}{
\begin{tabular}{l|cccccccc|c|c}
\specialrule{1pt}{0.5pt}{0.5pt}
& cables1 & camshake1 & einstein1 & plant1 & plant2 & sofa1 & table3 & table7 & Avg & FPS \\
\specialrule{1pt}{0.5pt}{0.5pt}
\spanner{}~\cite{WangX24Spanner3DReconstructionWithSpatialMemory} 
              & 33.2 & \textbf{5.1} & 30.9 & 4.1 & 5.7 & 17.1 & 19.3 & 18.9 & 16.8 & 6.1 \\ 
\munster{}-C    & 20.7&5.6&15.4&2.3&2.7&15.8&17.6& 9.5&11.2 & \textbf{11.8} \\
\munster{} &\textbf{20.7}&5.3&\textbf{11.2}&\textbf{1.8} & \textbf{2.7}& \textbf{15.5}&\textbf{17.3} &\textbf{5.5} & \textbf{10.0} & 8.7 \\
\specialrule{1pt}{0.5pt}{0.5pt}
\end{tabular}}
\normalsize
\caption{\textbf{ETH3D SLAM~\cite{eth3d_slam} Tracking Accuracy} ATE RMSE [cm] on ETH3D benchmark. 
Dense unconstrained methods for 8 selected trajectories.}
\label{tab:eth3d_ate}
\end{center}
\vspace{-3mm}
\end{table}

\begin{table*}
\begin{center}
\renewcommand\arraystretch{1.2}
\setlength{\tabcolsep}{1pt} 
\tiny
\hspace{-3mm}
\resizebox{\textwidth}{!}{
\begin{tabular}{l|rlrlrlc|rlrlrlc|rlrlrlc|c|c}
\specialrule{1pt}{0.5pt}{0.5pt}
{\multirow{3}{*}{Methods}} &  \multicolumn{6}{c}{7 Scenes} & &\multicolumn{6}{c}{NRGBD}
 & &\multicolumn{6}{c}{DTU} & & \\
&  
\multicolumn{2}{c}{Acc $\downarrow$} 
&   \multicolumn{2}{c}{Comp $\downarrow$} 
&   \multicolumn{2}{c}{NC $\uparrow$}
&  & \multicolumn{2}{c}{Acc $\downarrow$} &
  \multicolumn{2}{c}{Comp $\downarrow$} & 
  \multicolumn{2}{c}{NC $\uparrow$} 
   & & \multicolumn{2}{c}{Acc $\downarrow$} &
  \multicolumn{2}{c}{Comp $\downarrow$} & 
  \multicolumn{2}{c}{NC $\uparrow$} 
& & FPS  & Mem \\
 &  Mean & Med. & Mean & Med. & Mean & Med. & & Mean & Med. & Mean & Med. & Mean & Med.
  & & Mean & Med. & Mean & Med. & Mean & Med. & & &\\
\specialrule{1pt}{0.5pt}{0.5pt}

F-Recon \cite{xu2023frozenrecon}  & 0.124 &  0.076 &  0.055 & 0.023 & 0.619 & 0.688 & &0.285 & 0.206 & 0.151 &  0.063 &  0.655 &  0.758 && - & - & - & - & - & - & &($\leq 1$) & \\
\duster{}-224 \cite{duster} & 0.029 & 0.012 &  0.028 & \textbf{0.009} & \textbf{0.668} & \textbf{0.768} 
& & 0.054 & 0.025 & 0.032 & 0.010 & \textbf{0.802} &  \textbf{0.953} 
& & \textbf{2.296} & \textbf{1.297} & 2.158 & 1.002 & \textbf{0.747} & \textbf{0.848}
&  & 0.74 (0.78) & 38.1G\\
\spanner{} \cite{WangX24Spanner3DReconstructionWithSpatialMemory} & 0.034 &  0.015 &  \textbf{0.024} & \textbf{0.009} & 0.664 & 0.763 & & 0.069 & 0.032 & 0.029 &  0.011 & 0.778 &  0.937 & 
& 4.785 & 2.268 & 2.743 & 1.295 &  0.721 & 0.823
& &  27.38 (65.49) & 5.0G\\
\specialrule{0.5pt}{0.5pt}{0.5pt}

\munster{}-224 & 0.028 & 0.012 & 0.027 & 0.010 & 0.665 & 0.758 & & 0.062 & 0.025 & 0.031 & 0.012 & 0.788 & 0.930 & & 3.256 & 1.863 & 2.193 & 0.995 & 0.715 & 0.815 & & \textbf{40.41} & \textbf{4.1G} \\

\specialrule{1pt}{0.5pt}{0.5pt}

\munster{}-512 & \textbf{0.026} & \textbf{0.009} &  0.027 & \textbf{0.009} & 0.617 & 0.682 & & \textbf{0.048} & \textbf{0.022} & \textbf{0.020} & \textbf{0.008} & 0.768 &  0.911 & & 3.261 & 1.681 & \textbf{1.965} & \textbf{0.765} & 0.661 & 0.741 & & 12.10 & 8.1G \\

\specialrule{1pt}{0.5pt}{0.5pt}
\end{tabular}}
\normalsize
\caption{
\textbf{Comparison with \spanner{}}. FPS numbers in parenthesis are from \cite{WangX24Spanner3DReconstructionWithSpatialMemory}. Ours were obtained on a A100 GPU. We limit the maximum batch size when \emph{rendering} to 10 for \munster{}-224 and 5 for \munster{}-512. For \duster{}-224, our FPS and GPU memory numbers were obtained with a complete graph.}
\label{tab:spannercompar}
\end{center}
\end{table*}

\subsection{Uncalibrated Visual Odometry}
\label{ssec:vo}
We evaluate our online \munster{} model on the real images datasets TUM RGBD~\cite{Sturm2012ASystems} and ETH3D SLAM~\cite{eth3d_bench, eth3d_slam}. We cannot compare on Replica~\cite{straub2019replica} since it is contained in our training set. Note that contrary to the entirety of the compared methods, \munster{} and \spanner{} operate in the unconstrained scenario, meaning the calibration of the camera system is considered unknown. 
Our results are obtained with the same hyperparameters for all datasets $\tau_d=5\%$ and $p_d=85\%$ (see Tab.~\ref{sec:memory}). We denote \munster{}-C the causal variant, without \emph{rendering}, while \munster{} with \emph{rendering} is followed by a minimal Laplacian smoothing. We compare to \spanner{} by running their code on the SLAM benchmarks, with minimal modification to recover camera poses via PnP on the predictions, leveraging the focal length estimated by the first view.

We are interested in the RMSE Average Trajectory Error (ATE) after alignment to the \emph{ground-truth} (GT) as a measurement of camera trajectory quality. For unconstrained methods, we also look at the frame-rate, absolute vertical Field-of-View (FoV) error in degrees, as well as metric scale estimation quality in \% of the ground-truth scale. Following prior art~\cite{DexheimerCVPR23PCOMOCompactMappingOdometry,teed_dpvo_2023}, for fair comparison, all SLAM methods run in visual odometry (VO) mode, \ie without global bundle adjustment. Qualitative results of the trajectory and reconstructions for both datasets are showed in Fig.~\ref{fig:qualitative}, \ref{fig:overview} and \ref{fig:qualitative1_supmat}. 

\noindent \textbf{TUM RGBD} \cite{Sturm2012ASystems} is a well established benchmark that presents challenges for monocular VO and SLAM due to significant motion blur, exposure changes, large rotations and trajectories, dynamic scenes and rolling shutter artifacts. We compare our method against state-of-the-art methods on 11 RGB-only sequences (Tab.~\ref{tab:tum_ate}) and we report all sequences in the appendix in Tab.~\ref{tab:tum_full}. 

\munster{} and \spanner{}~\cite{WangX24Spanner3DReconstructionWithSpatialMemory} form a new group of dense unconstrained methods (U). As the table shows, \spanner{} is penalized by high error on four long sequences. Instead, \munster{}-C performs well on both short and long sequences. Adding \emph{rendering} further reduces the error and achieves the best performance on average at the cost of a slightly decreased frame rate ($8.4$ FPS). For these three methods, table reports the average speed (FPS) on NVIDIA V100 GPU.
For the (U) category, we can evaluate the accuracy of the focal estimates in terms of absolute FoV error. Tab.~\ref{tab:tum_fov} compares these values for \munster{} and \spanner{} on the same subset of 8 trajectories, with a clear advantage of the former, achieving an average error of $4^\circ$ only. Tab.~\ref{tab:tum_scale} shows that \munster{}-C and \munster{} estimate scale with a comparable accuracy, with a median scale error of $5.5\%$ and $4.6\%$ respectively.

Tab.~\ref{tab:tum_ate} contains the $11$ sequences commonly used for evaluating in the VO setting \cite{teed21droid-slam,DexheimerCVPR23PCOMOCompactMappingOdometry,ZhangX24GlORIESLAM}. We complete the evaluation by running both \spanner{} and \munster{} on the full set of 46 test scenes.
We point out that many of these scenes represent a real challenge for current state-of-the-art methods. We find that \munster{} compares favorably (Tab.~\ref{tab:tum_group}) on all six categories as defined by the challenge: Testing, Handheld, Robotic, Structure vs Texture, Dynamic Objects and 3D Objects Recognition. 

\noindent \textbf{ETH3D SLAM}~\cite{eth3d_bench} includes a large number of 
trajectories recorded in a motion capturing system. \munster{} is $50\%$ more accurate on average than \spanner{} (Tab.~\ref{tab:eth3d_ate}; see Tab.~\ref{tab:eth3d_full} in the appendix\ for the full set of results) with again an average RMSE of $10.0$ cm, obtained at around $10$ FPS.

\subsection{Relative pose estimation}
\label{ssec:relpose}

We tested \munster{} and \spanner{} on the CO3D~\cite{co3d} and RealEstate10k~\cite{ZhouTOG18Realestate10K} datasets, which are indoor/outdoor datasets where camera poses were obtained via COLMAP or SLAM with bundle adjustment on the full image sequences. 
As~\cite{posediffusion}, we evaluate the methods on the 1.8K video clips from the test set.  Each sequence is 10 frames long and we evaluate relative camera poses between all possible 45 pairs. As defined in \cite{imc22, ZhangICLR24CamerasasRaysPoseEstimationViaRayDiffusion, relpose}, reported metrics are Relative Rotation Accuracy below $15^{\circ}$ (RRA@15), Relative Translation direction Accuracy below $15^{\circ}$ (RTA@15) and mean Average Accuracy below $30^{\circ}$ (mAA@30) defined as the area under the curve $min$(RRA@$\tau$, RTA@$\tau$) at a threshold $\tau$, integrated over $[1,30]$.
\begin{table}[t]
    \begin{center}
\renewcommand\arraystretch{1.2}
\setlength{\tabcolsep}{1pt} 
\tiny
\resizebox{\columnwidth}{!}{
\begin{tabular}{l|ccc|cc|c}
\specialrule{1pt}{0.5pt}{0.5pt}
 \multirow{2}{*}{Method}  & \multicolumn{3}{c|}{Co3Dv2$\uparrow$} &  & RealEstate10K$\uparrow$ & Speed \\ \cline{2-4} \cline{6-6} 
 \hspace{0.1pt} & RRA@15  & RTA@15 & mAA(30) &  & mAA(30)   &   (FPS)  \\ 
\specialrule{0.5pt}{0.5pt}{0.5pt}
\duster-512~\cite{duster}               & 94.3    & 88.4  & 77.2    &  & 61.2       &   3.2 \\
\duster-512-GA~\cite{duster}             & 96.2    & 86.8   & 76.7    &  & 67.7     &  0.1   \\ 
Spann3R~\cite{WangX24Spanner3DReconstructionWithSpatialMemory} &  89.1   & 83.6 & 70.4  &  &  60.8      &  7.4 \\
\munster -224 & 95.1 & 90.8 & 80.7 &  & 74.7& 11.7 \\    
\munster -512 & \textbf{97.0} &\textbf{ 92.7} & \textbf{84.1} &  & \textbf{75.1}& 4.1 \\  
\munster -512 (Pro) & 95.5 & 88.9& 78.3 &  & 65.5 & \textbf{32.9} \\  
 
\specialrule{1pt}{0.5pt}{0.5pt}
\end{tabular}}
    \caption{\textbf{Multi-view pose regression on CO3Dv2~\cite{co3d} and 
            RealEstate10K~\cite{ZhouTOG18Realestate10K} with 10 random frames}. All methods resort to PnP to estimate camera pose except for (Pro) that uses Procrustes alignment. FPS include both inference and pose estimation, on a A100 GPU.
       }
    \label{tab:relpose_mvs}
    \end{center}
\end{table}

We compare \munster{} to previous works based on pointmap regression. If not specified otherwise, the focal length is estimated from the predictions, and later used with PnP~\cite{hartleymultiviewgeometry} to recover the camera poses. In the case of \duster{}, all pairs have to be processed twice through the decoder. The Global Alignment (GA) adds another layer of computational complexity to align all pairwise predictions in the same reference frame. In contrast, \spanner{} leverages a memory mechanism similar in spirit to ours, decreasing the computational complexity and increasing the FPS at the cost of accuracy. 
\munster{}, which also leverages a working memory expressing all views in the same coordinate system, yields better performance than all baselines. For fairness, \spanner{} needs to be compared to the 224 resolution where \duster{} is competing against the 512 one.
The (Pro) variant is the one that leverages Procrustes alignment for pose estimation. It provides almost the same accuracy than PnP but is almost an order of magnitude faster, which is of critical importance for real-time applications, like Sec.~\ref{ssec:vo}.

\subsection{3D Reconstruction}
\label{ssec:3dr}
We evaluate pointmaps on 7Scenes~\cite{7scenes}, Neural RGBD~\cite{AzinovicCVPR22NeuralRGBDSurfaceReconstruction} and DTU~\cite{AanaesIJCV16dtu} using the same protocol as \spanner{} \cite{WangX24Spanner3DReconstructionWithSpatialMemory}. In more details, Following prior works \cite{WangCVPR23CoSLAMJointNeuralRealTimeSLAM,ZhuCVPR22NICESLAMNeuralImplicitScalableEncoding4SLAM,WangX24Spanner3DReconstructionWithSpatialMemory}, we report in Tab.~\ref{tab:spannercompar} accuracy (Acc), completeness (Comp) and normal consistency (NC) such that the predicted dense pointmap is directly compared with the back-projected per point depth, excluding invalid and background points if applicable. For \munster{}, we update the memory with all images, and then \emph{render} the final pointmaps. 

Quantitatively, \munster{} almost always outperforms \spanner{} with better FPS. 
It also achieves a performance similar to \duster{}, while being $5$ times lighter and an order of magnitude faster. We note that the GPU memory usage between \spanner{} and \munster{} is in the same range, the small improvement of \munster{} could be attributed to the difference in implementations, esp. in the attention. 
\munster{}-224 uses 4.1G with xformers and 5.2G with the naive implementation, close to the 5.0G of \spanner{}. Detailed studies on the impact of the memory size and the number of images passed at once are available in Sec.~\ref{sec:runtime}.

\begin{table}
\begin{center}
\renewcommand\arraystretch{1.2}
\setlength{\tabcolsep}{1pt} 
\tiny
\hspace{-3mm}
\resizebox{\columnwidth}{!}{
\begin{tabular}{l|rrrrrrrrrr|rrc}
\specialrule{1pt}{0.5pt}{0.5pt}
\multirow{2}{*}{Methods} & \multicolumn{2}{c}{KITTI} & \multicolumn{2}{c}{ScanNet} & \multicolumn{2}{c}{ETH3D} & \multicolumn{2}{c}{DTU} & \multicolumn{2}{c}{T\&T} & \multicolumn{3}{|c}{Avg} \\
 & rel $\downarrow$ & $\tau \uparrow$ & rel $\downarrow$ & $\tau \uparrow$ & rel $\downarrow$ & $\tau \uparrow$ & rel $\downarrow$ & $\tau \uparrow$ & rel$\downarrow$ & $\tau \uparrow$ & rel$\downarrow$ & $\tau \uparrow$ & time (s)$\downarrow$ \\
\specialrule{1pt}{0.5pt}{0.5pt}

{\duster}-512 ~\cite{duster}& 5.4 & 49.5 & (\textbf{3.1}) &  (\textbf{71.8}) & 3.0 & 76.0 & \textbf{3.9} & \textbf{68.6} & 3.3 & 75.1 & \textbf{3.7}& \textbf{68.2} & 0.19 \\
{\munster}-512 & \textbf{4.5} & \textbf{55.0} & (4.0)&(59.8) & \textbf{2.5} & \textbf{80.3} & 4.6 & 55.4 &(\textbf{2.6})&(\textbf{80.4})& \textbf{3.7} & 66.2 & 0.29 \\

\specialrule{0.5pt}{0.5pt}{0.5pt}

{\duster}-224~\cite{duster}&  9.2 & 32.9 & (4.2) & (58.2) & 4.7 & 61.9 & \textbf{2.8}& \textbf{77.3} & 5.5 & 56.5 & 5.3 & 57.4 & 0.10 \\

\spanner{} ~\cite{WangX24Spanner3DReconstructionWithSpatialMemory}& 7.9 & 36.2 & (\textbf{3.3})&(\textbf{67.1}) & 5.7 & 58.6 & 3.5 & 65.2 &(\textbf{4.7})&(58.5)& 5.0 & 57.1 & 0.32 \\

 {\munster}-224 & \textbf{6.1} & \textbf{46.8} & (4.5)&(56.7) & \textbf{3.6} & \textbf{68.0} & 4.6 & 63.1 &(\textbf{4.7})&(\textbf{64.5})& \textbf{4.7} & \textbf{59.8} & 0.19 \\
 
\specialrule{1pt}{0.5pt}{0.5pt}
\end{tabular}}

\normalsize
\caption{
\textbf{Multi-view depth evaluation} with no poses nor intrinsics. Results are for the quasi-optimal number of compared views $n \in [1,10] $ for each method.
(Parentheses) denote training on data from the same domain. Best results  
(per resolution) in \textbf{bold}.
}
\label{tab:mvd_supp}
\end{center}
\end{table}

\subsection{Multi-view depth evaluation}
\label{ssec:mvd}
Finally, we compare \munster{} to pointmap regression methods on multi-view stereo depth estimation. In our case, depthmaps are simply the $z$-coordinate of the local pointmaps $X_{i,i}$.
As~\cite{robust_mvd}, we evaluate it on the KITTI~\cite{kitti}, ScanNet~\cite{scannet}, ETH3D~\cite{eth3d}, DTU~\cite{AanaesIJCV16dtu} and Tanks and Temples (T\&T)~\cite{tandt} datasets, reporting Absolute Relative Error (rel) and Inlier Ratio $(\tau)$ with a threshold of 1.03 on each test set, and the averages across all test sets. Because none of the methods leverage \emph{ground-truth} camera parameters and poses, the predictions have to be aligned to \emph{ground-truth} via median normalization as advocated in~\cite{robust_mvd}. 
On average \munster{}-$224$ performs better than other baselines and the $512$ version performs similarly to \duster{}. 

\subsection{Limitations}
Despite very strong results on multiple downstream tasks, \munster{} shows signs of limitations for sequences where the views drift too far from the 1st view.

\section{Conclusion}
\label{sec:concusion}
We proposed \munster{}, a new multi-view network for 3D reconstruction of large image collections which operates in offline and online scenarios at high speed. 
Our evaluations shows the state-of-the-art performance of \munster{} on multiple 3D downstream tasks, such as depth and relative pose estimation, 3D reconstruction and uncalibrated VO.

\clearpage  
\appendix

{\large \textbf{Appendix}}

This appendix first includes in Appendix~\ref{supsec:ablation} all ablation studies that justify the design choices of the~\munster{} architecture. We then provide the promised qualitative examples of~\munster{} on real scenes from various datasets in Appendix~\ref{supsec:quali} and the snippet of code for the online reconstruction in Appendix~\ref{supsec:code}. Finally, we detail in Appendix~\ref{supsec:vo} the complete quantitative evaluations of our method on the full sets of sequences from TUM-RGBD\footnote{https://cvg.cit.tum.de/data/datasets/rgbd-dataset/} and ETH3D\footnote{https://www.eth3d.net/slam\_datasets} datasets~\cite{Sturm2012ASystems,eth3d_slam}. 

\begin{figure}
    \centering
    \includegraphics[width=0.9\linewidth, trim=0 0 0 0, clip]{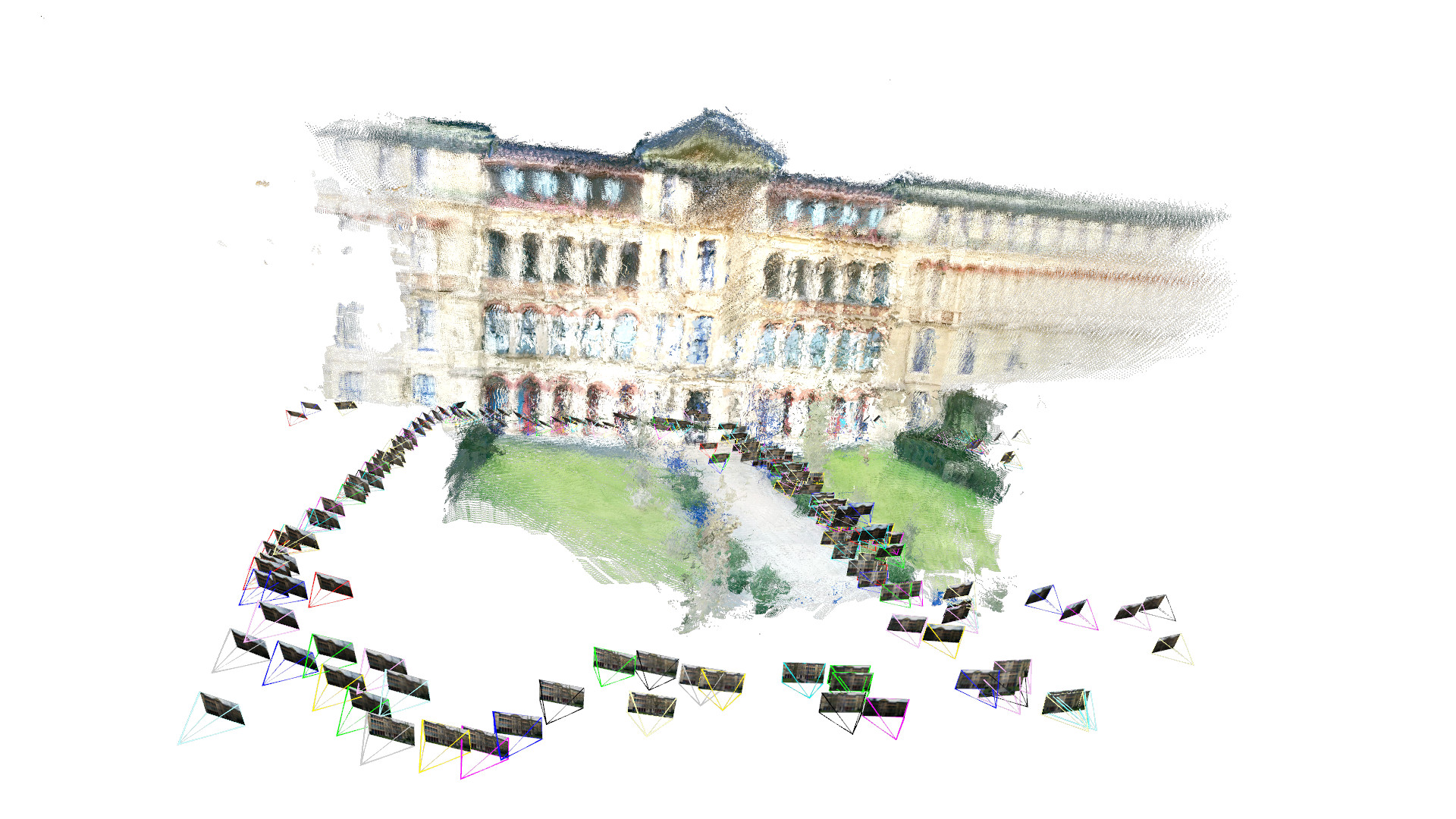}
    \includegraphics[width=0.9\linewidth, trim=0 0 0 0, clip]{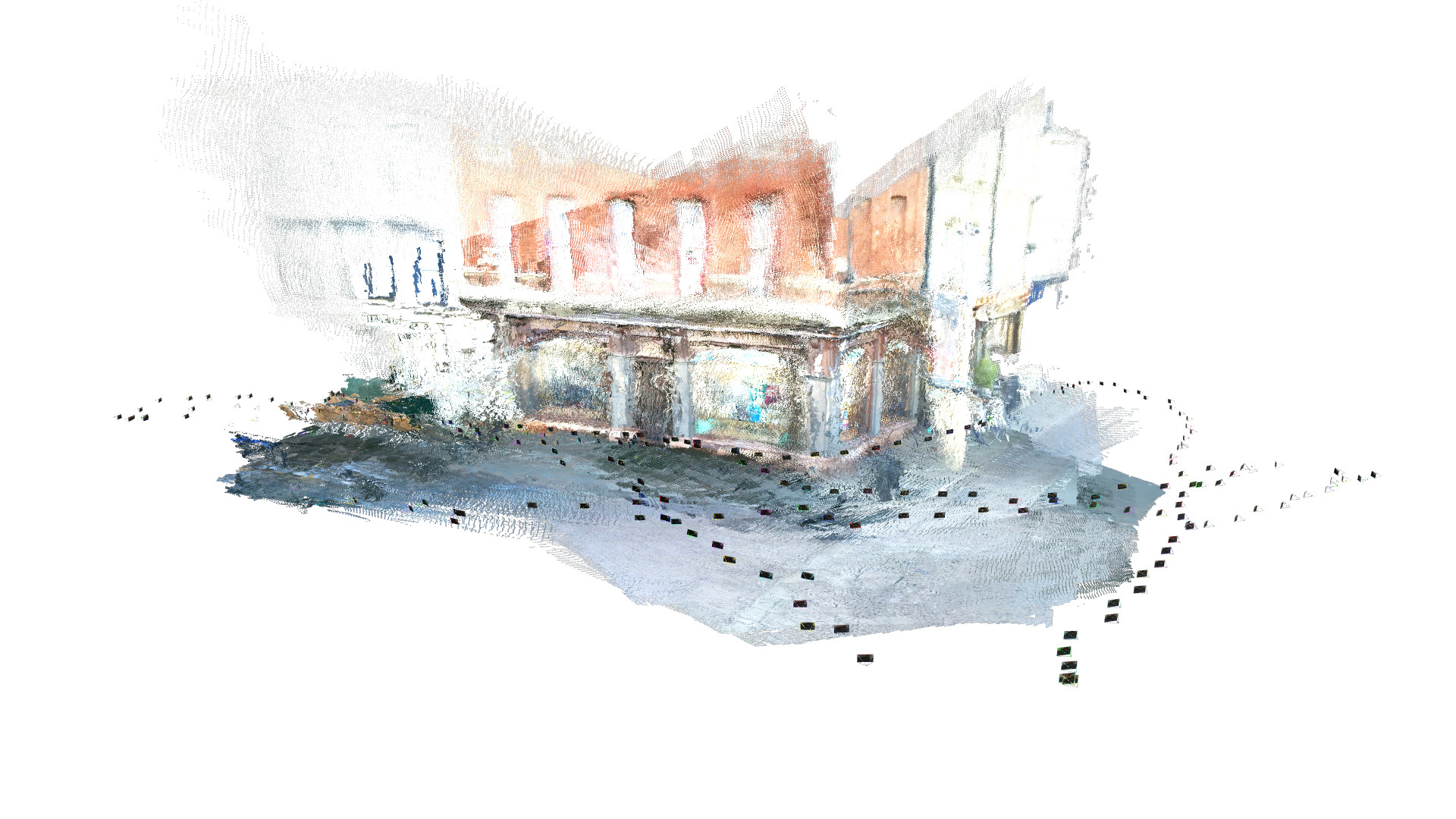}
    \includegraphics[width=0.9\linewidth, trim=0 0 0 0, clip]{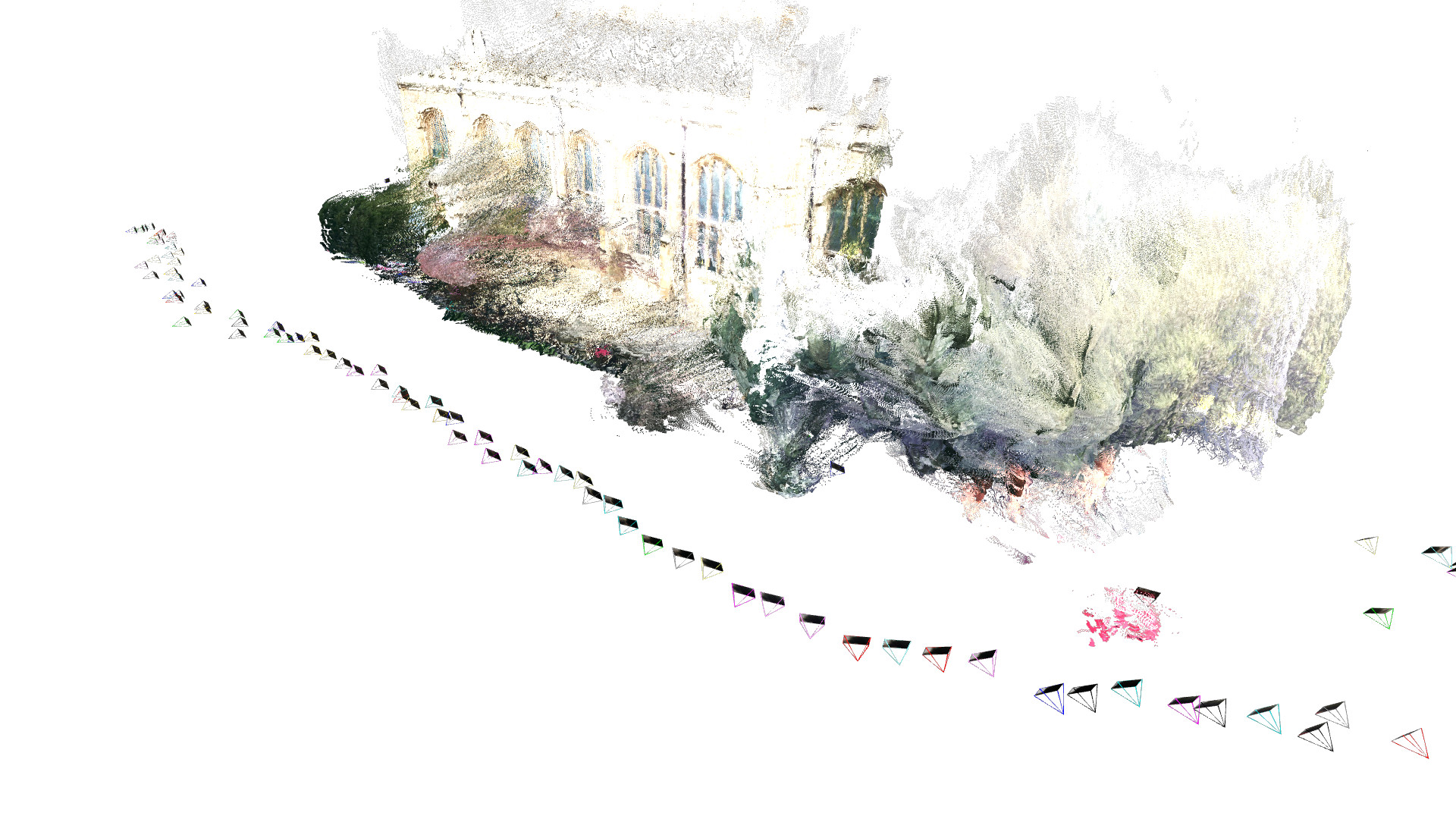}
      \includegraphics[width=0.9\linewidth, trim=0 0 0 0, clip]{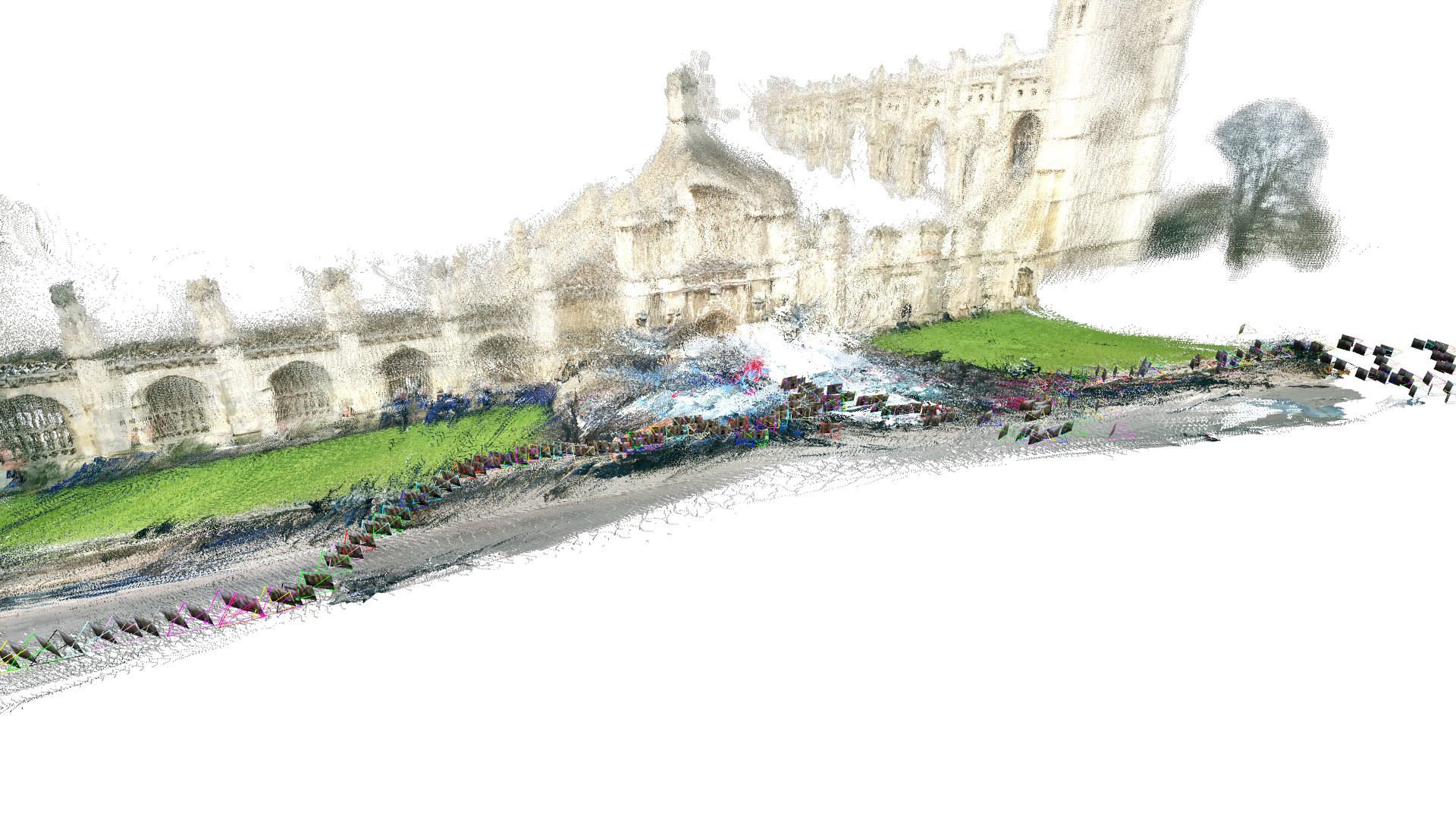} 
      \includegraphics[width=0.9\linewidth, trim=0 0 0 0, clip]{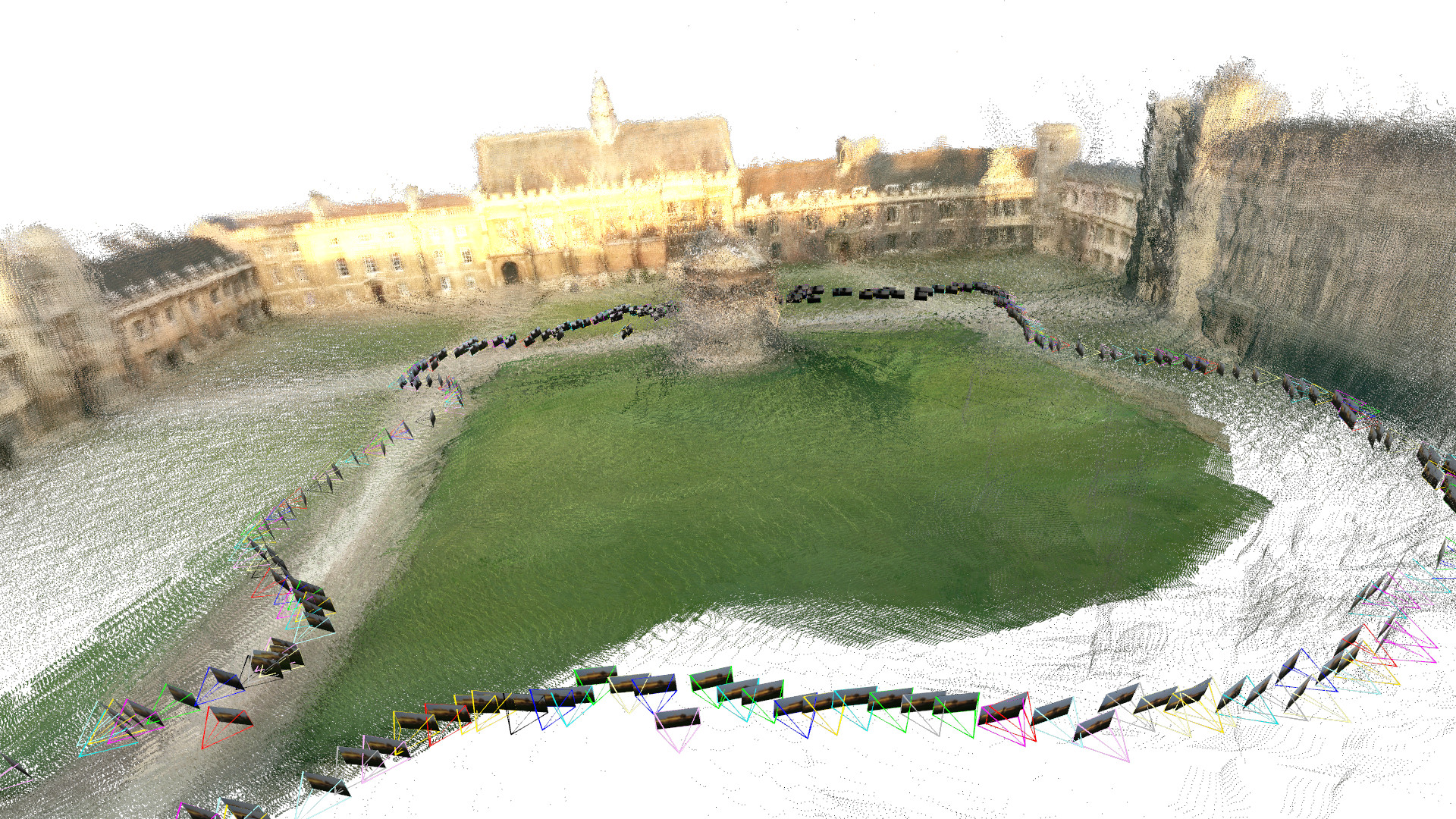}
    \caption{
        Qualitative example of \munster{} reconstructions of Cambridge Landmarks \cite{cambridge}.
    }
    \label{fig:qualitative3_supmat}
\end{figure}

\section{Ablation studies}
\label{supsec:ablation}

The aim of this section is to provide empirical analysis of our design choices, starting with our simplification of the \duster{} architecture in Appendix~\ref{supssec:symduster}, and our memory management and loss choices in Appendix~\ref{sec:archi}. Then in Appendix~\ref{sec:runtime} we study the scalability of the network to more views than seen during training.

\begin{figure*}[ttt]
    \centering   \includegraphics[width=0.32\linewidth, trim=0 0 0 0, clip]{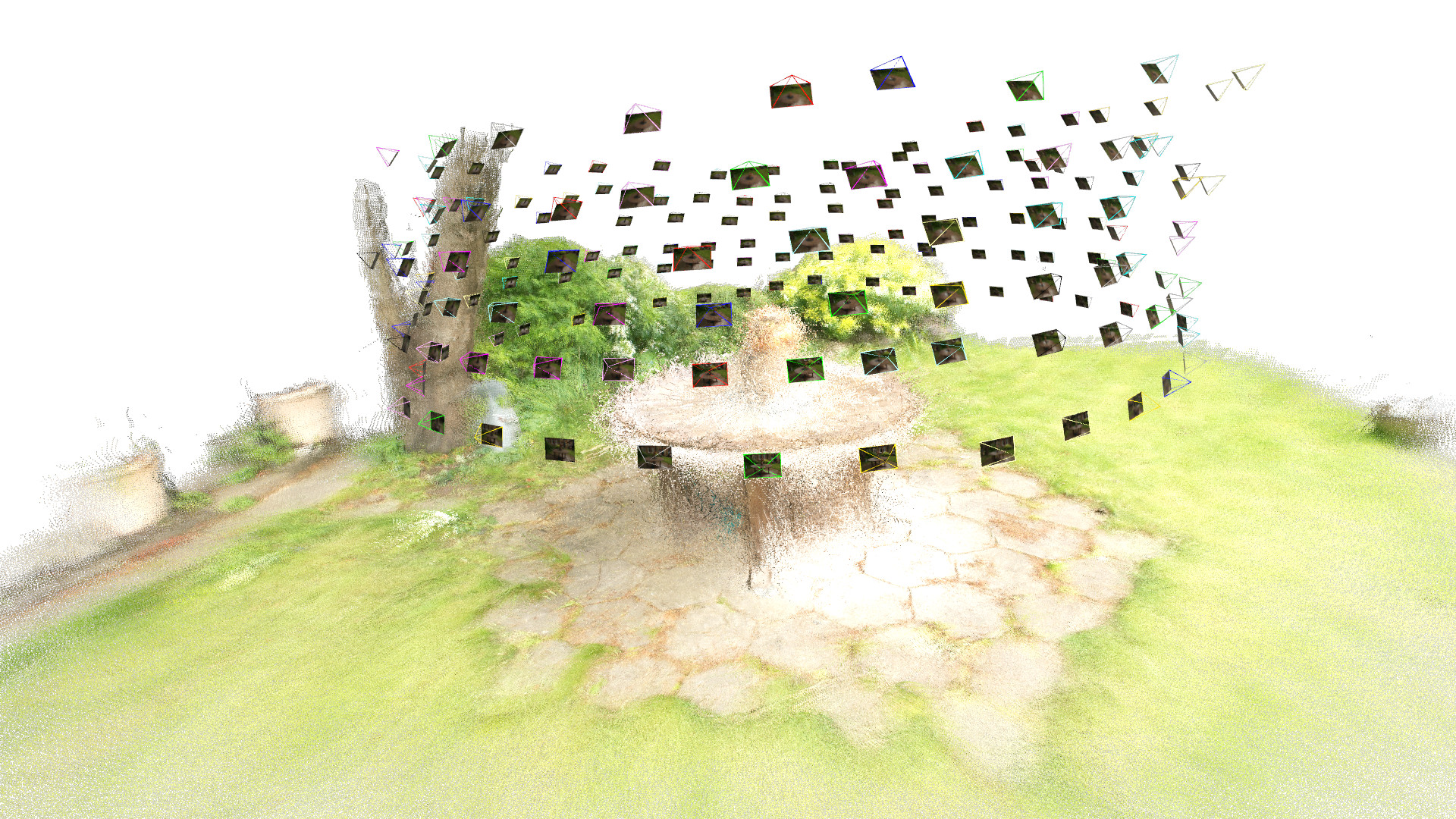}  \includegraphics[width=0.32\linewidth, trim=0 0 0 0, clip]{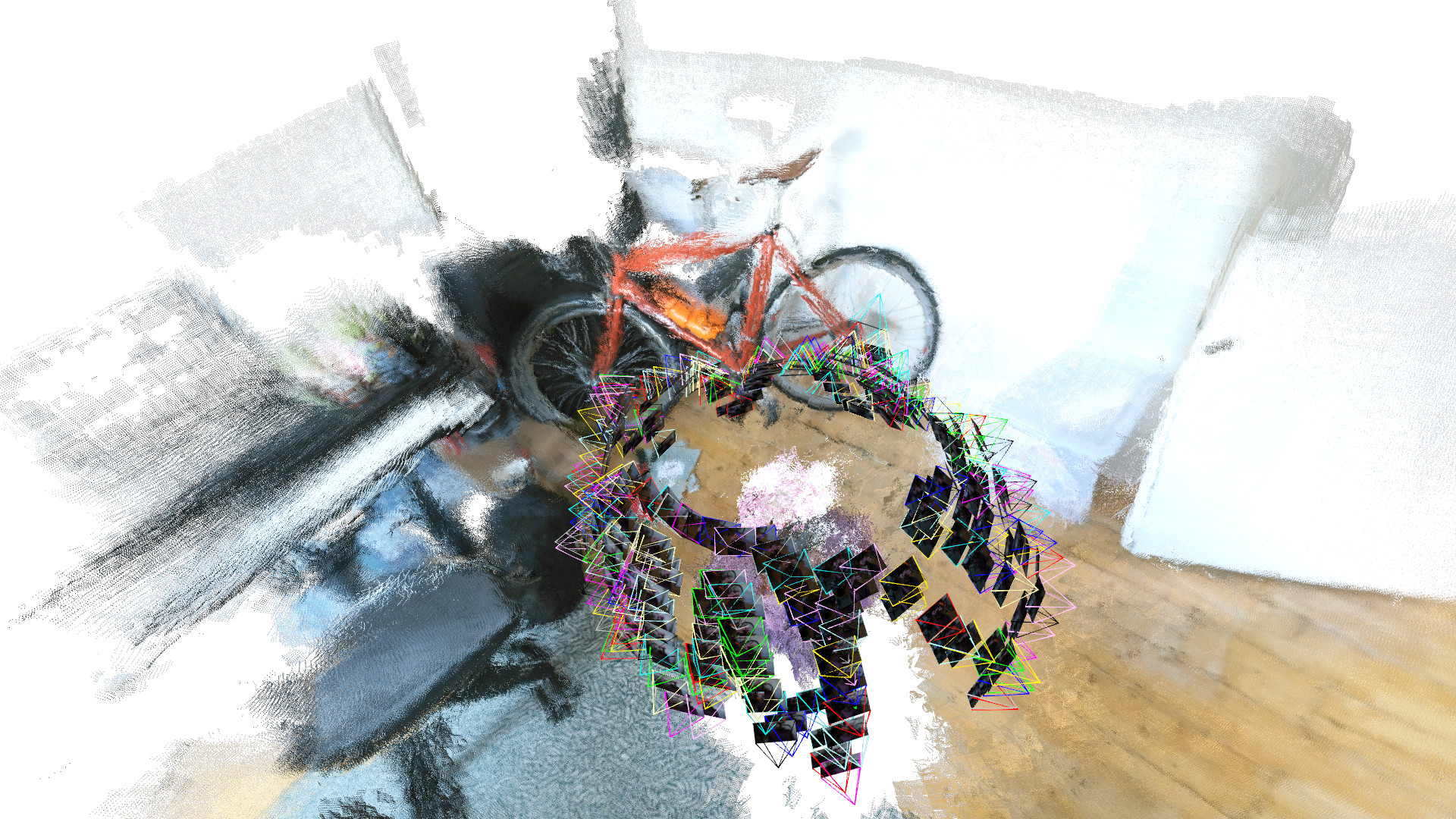}   \includegraphics[width=0.32\linewidth, trim=0 0 0 0, clip]{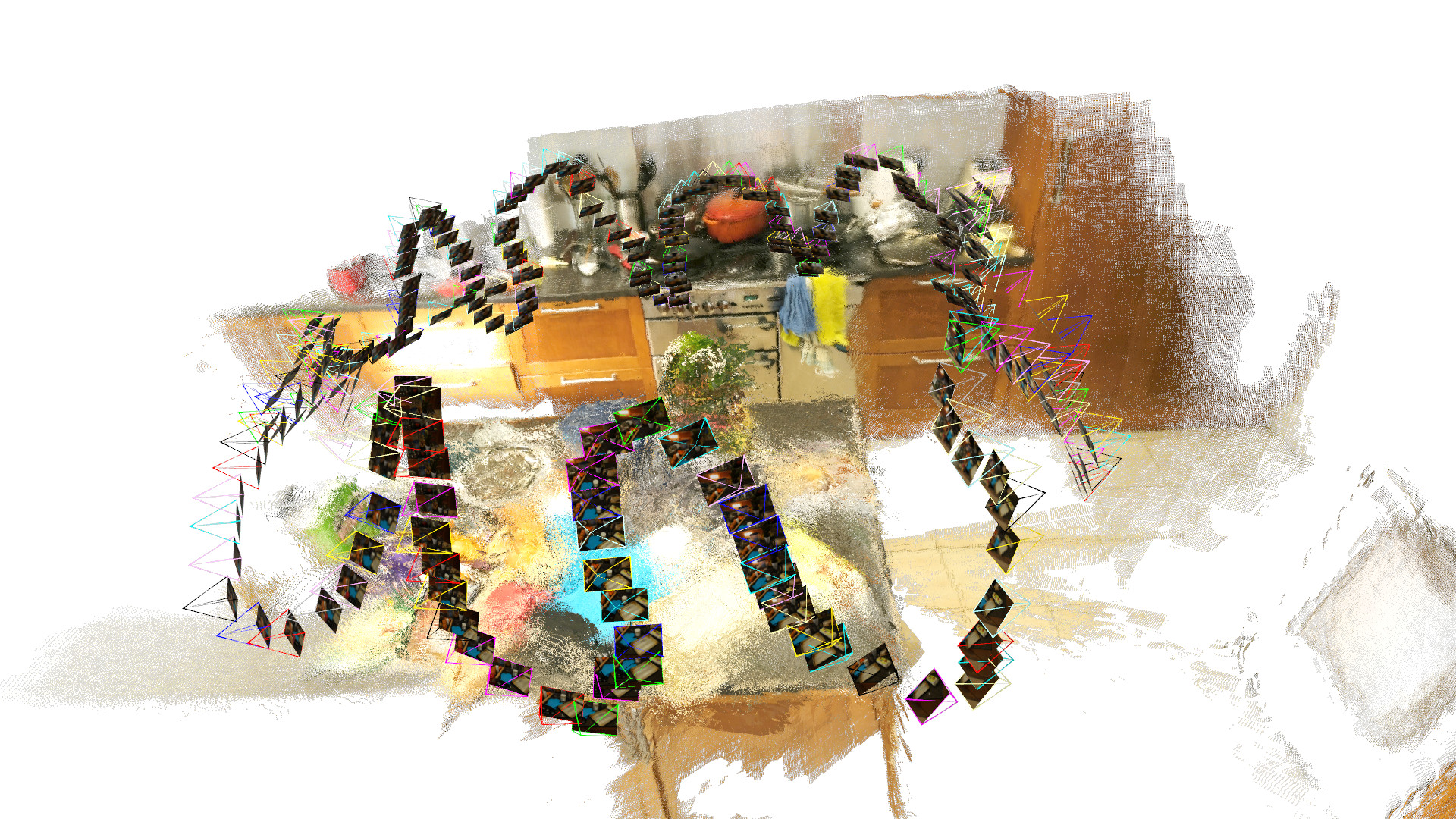}   \includegraphics[width=0.32\linewidth, trim=0 0 0 0, clip]{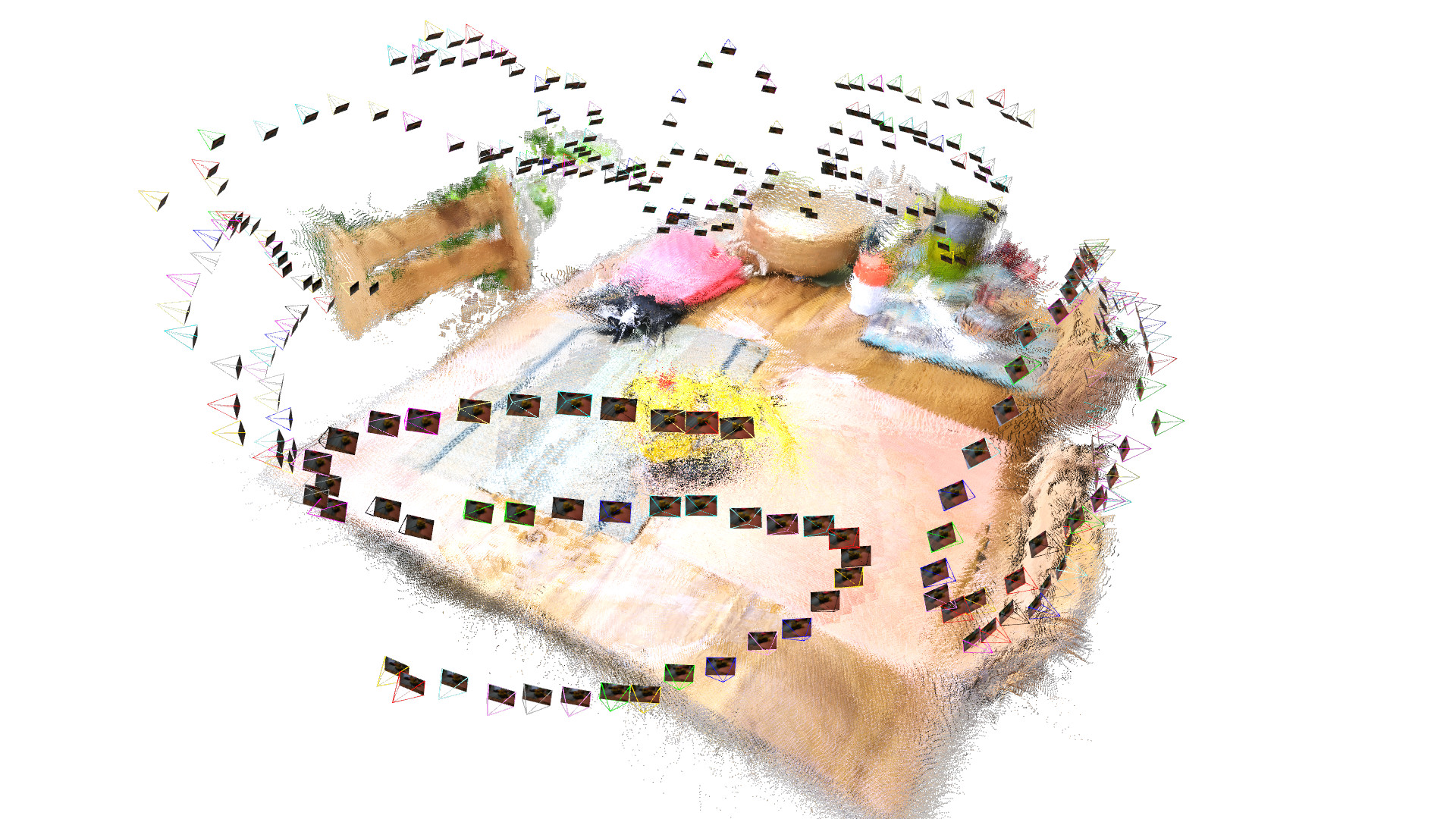}   \includegraphics[width=0.32\linewidth, trim=0 0 0 0, clip]{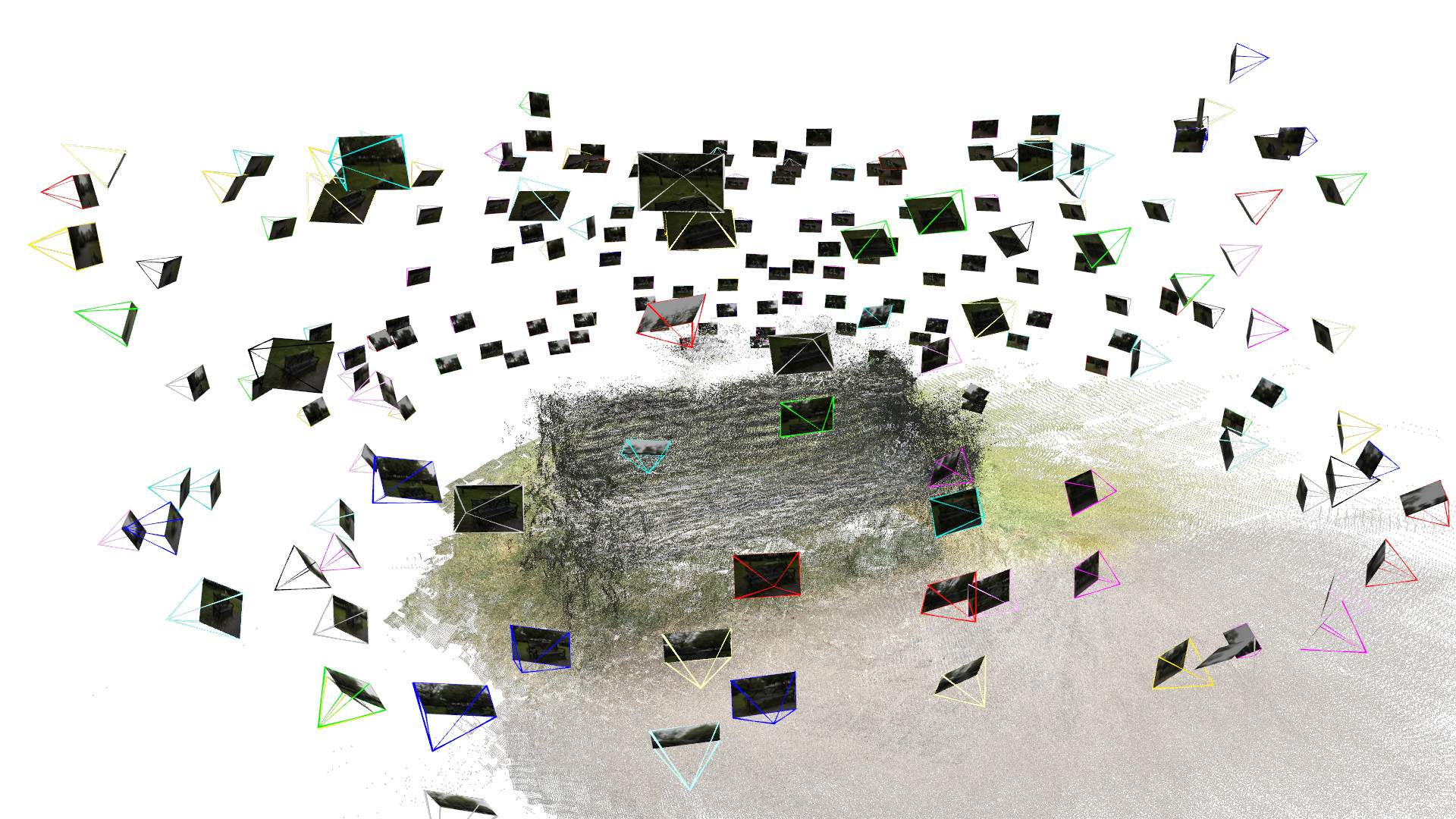} \includegraphics[width=0.32\linewidth, trim=0 0 0 0, clip]{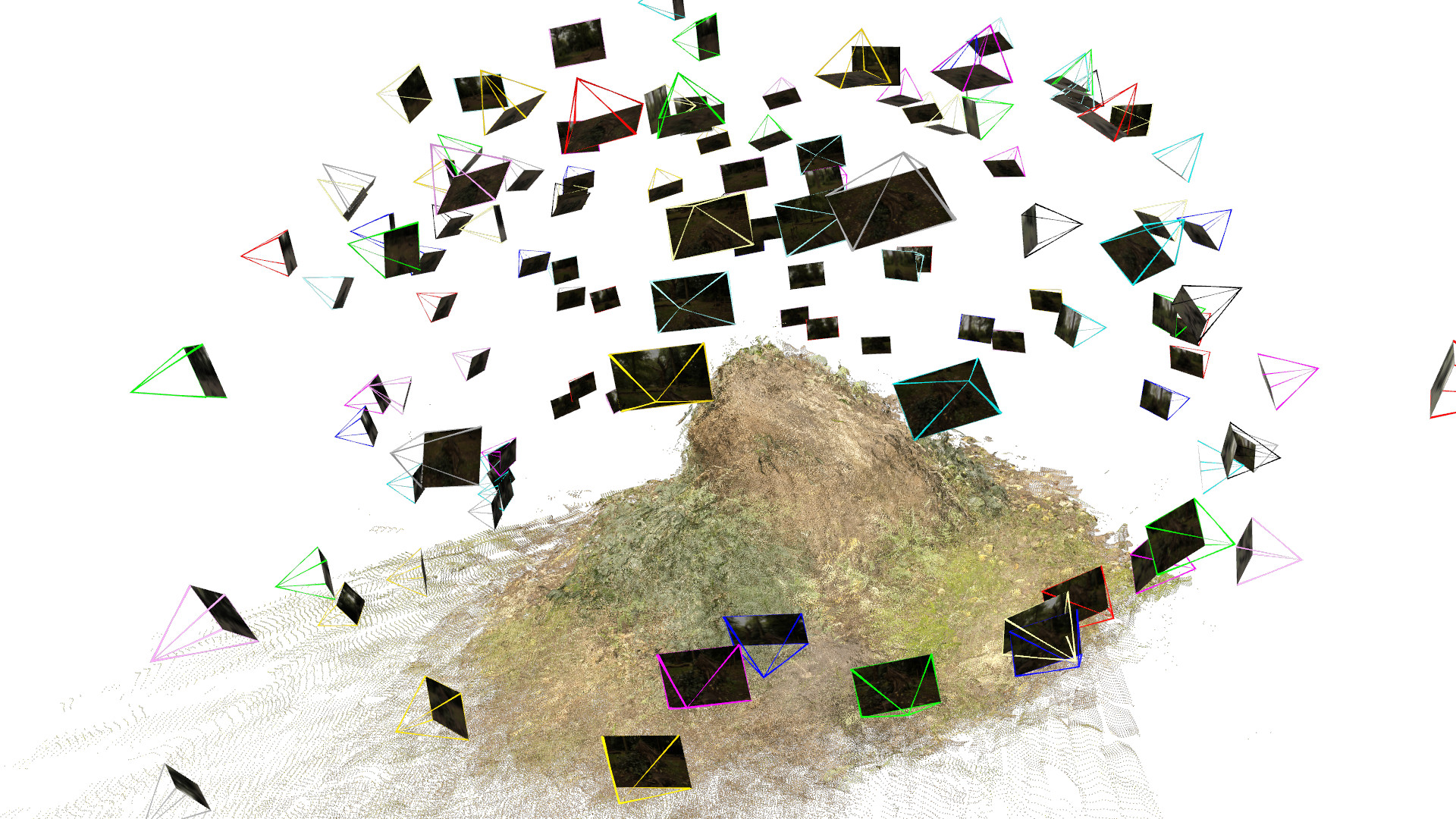}  \includegraphics[width=0.32\linewidth, trim=0 0 0 0, clip]{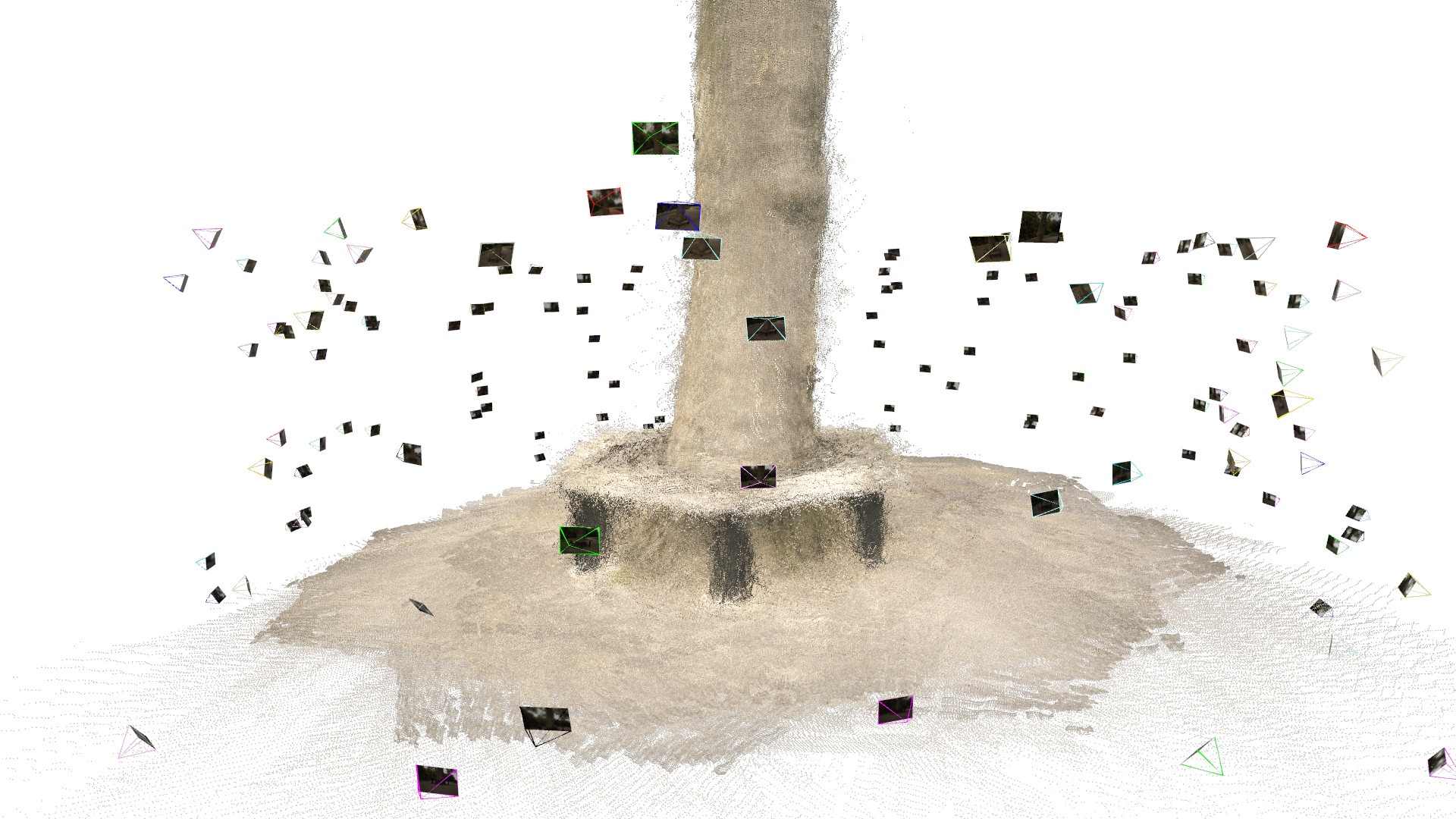}   \includegraphics[width=0.32\linewidth, trim=0 0 0 0, clip]{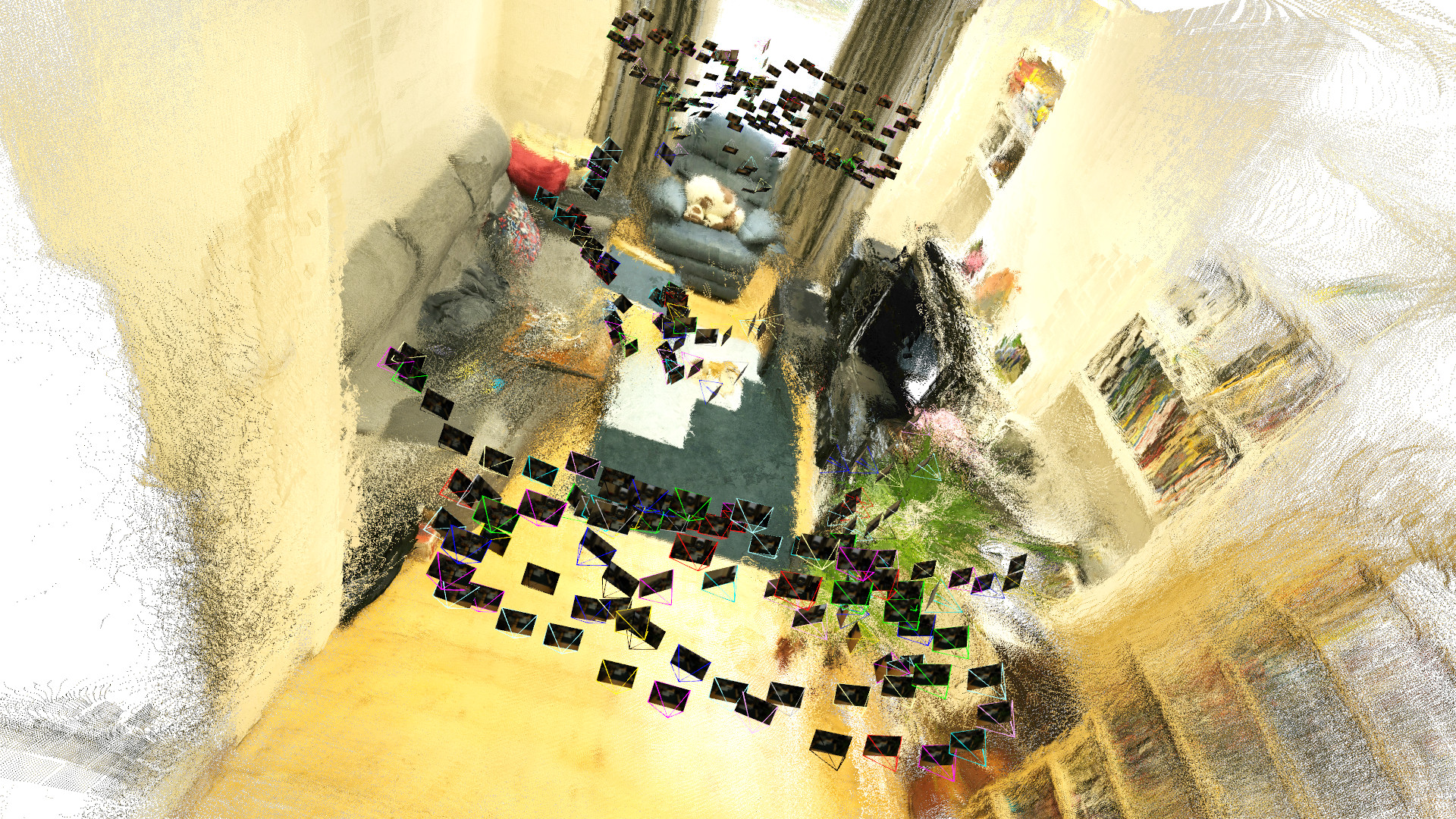}   \includegraphics[width=0.32\linewidth, trim=0 0 0 0, clip]{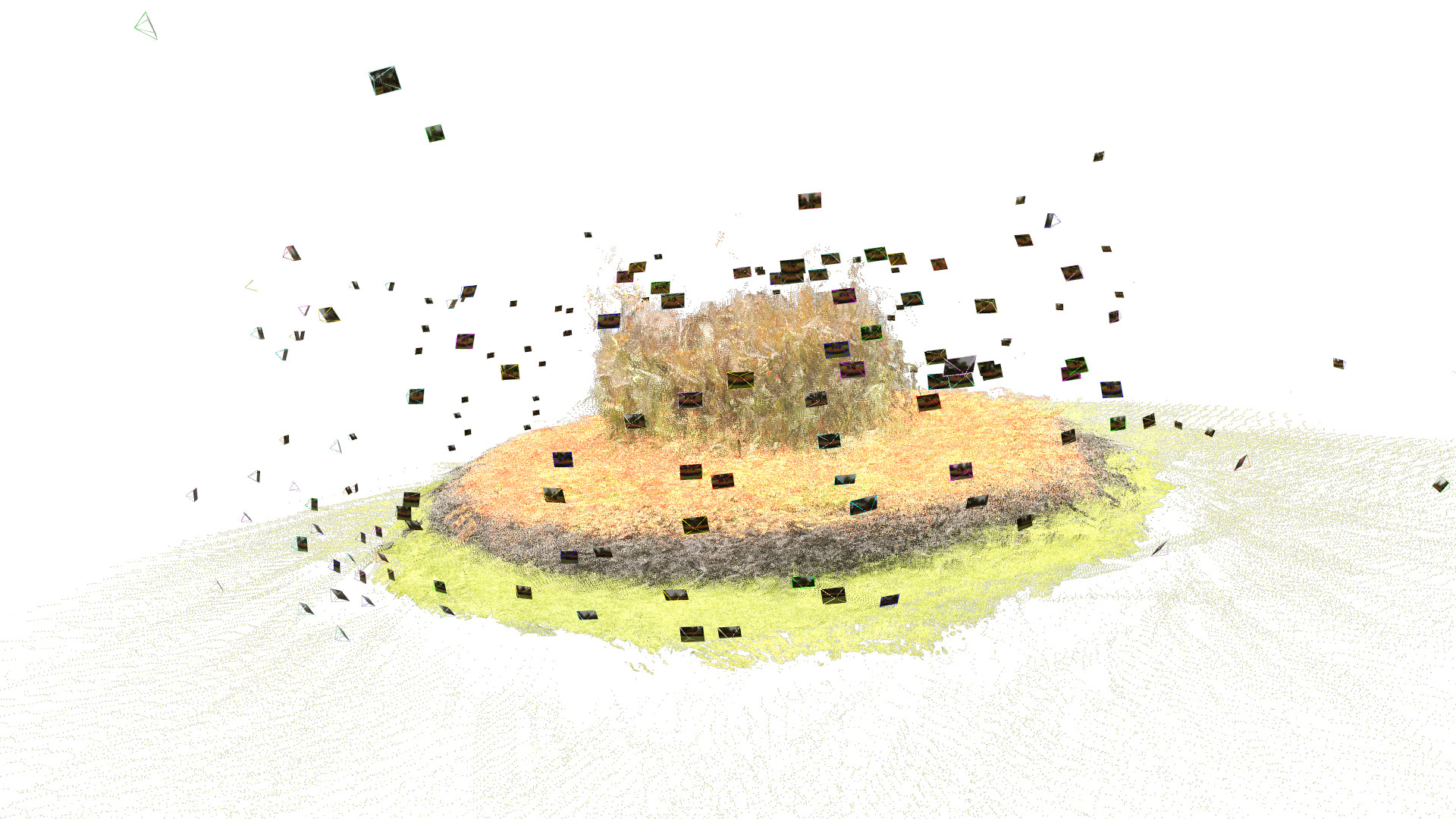}
    
    \caption{
        Qualitative example of \munster{} reconstructions of MIP-360 \cite{barron2022mipnerf360}.
    }
    \label{fig:qualitative2_supmat}
\end{figure*}

\subsection{Simplifying \duster}
\label{supssec:symduster}
As explained in the main paper, the architecture of \duster{} cannot easily scale to more views for its asymmetric design would require to train $N$ different decoders for $N$ views, which is not trivial and computationally impractical. For this reason, we propose to leverage a \emph{Siamese} decoder, \ie one decoder for all views, in a symmetric manner. The symmetric decoder is the core of our model as it enables to seamlessly scale to more views without the need to retrain the model. We evaluate here how this choice impacts the performance compared to the original \duster{} architecture. 
In more details, leveraging a symmetric decoder for $N$ views predictions requires to disable the Rotary positional encoding (RoPE) in the Cross-Attention (CA) and to add a learnable encoding to distinguish between the reference view, that defines the origin of the coordinate system, and all the other views. We ablate these in the following.

\mypar{Setup and metric.} To diminish compute and increase speed, we perform ablations in the pairwise setting ($N{=}2$) on a small subset of datasets namely the union of subsets of Habitat and Megadepth, for a total of $2M$ pairs of $224 \times 224$ resolution images.

We evaluate performance on the validation sets of these two datasets. We believe this setup to be representative enough for us to reasonably extend the conclusions to the full training set. 
 We report in Tab.~\ref{tab:ablation_sym} the average regression error that directly measures the quality of predicted pointmaps. We observe this metric generally represents well the performance on downstream tasks. In the table, \textbf{Baseline} refers to the asymmetric \duster{} architecture. In \textbf{No RoPE in CA}, we remove RoPE in the cross attentions. In \textbf{Symmetric}, we switch to the symmetric decoder setup where we remove the second decoder and $\head$ and we use instead the same shared decoder and head for both images.
 In \textbf{Symmetric(embed)}, we add the learned embedding $B$, while it is absent in \textbf{Symmetric(none)}.

\begin{figure*}[ttt]
    \centering   \includegraphics[width=0.32\linewidth, trim=0 0 0 0, clip]{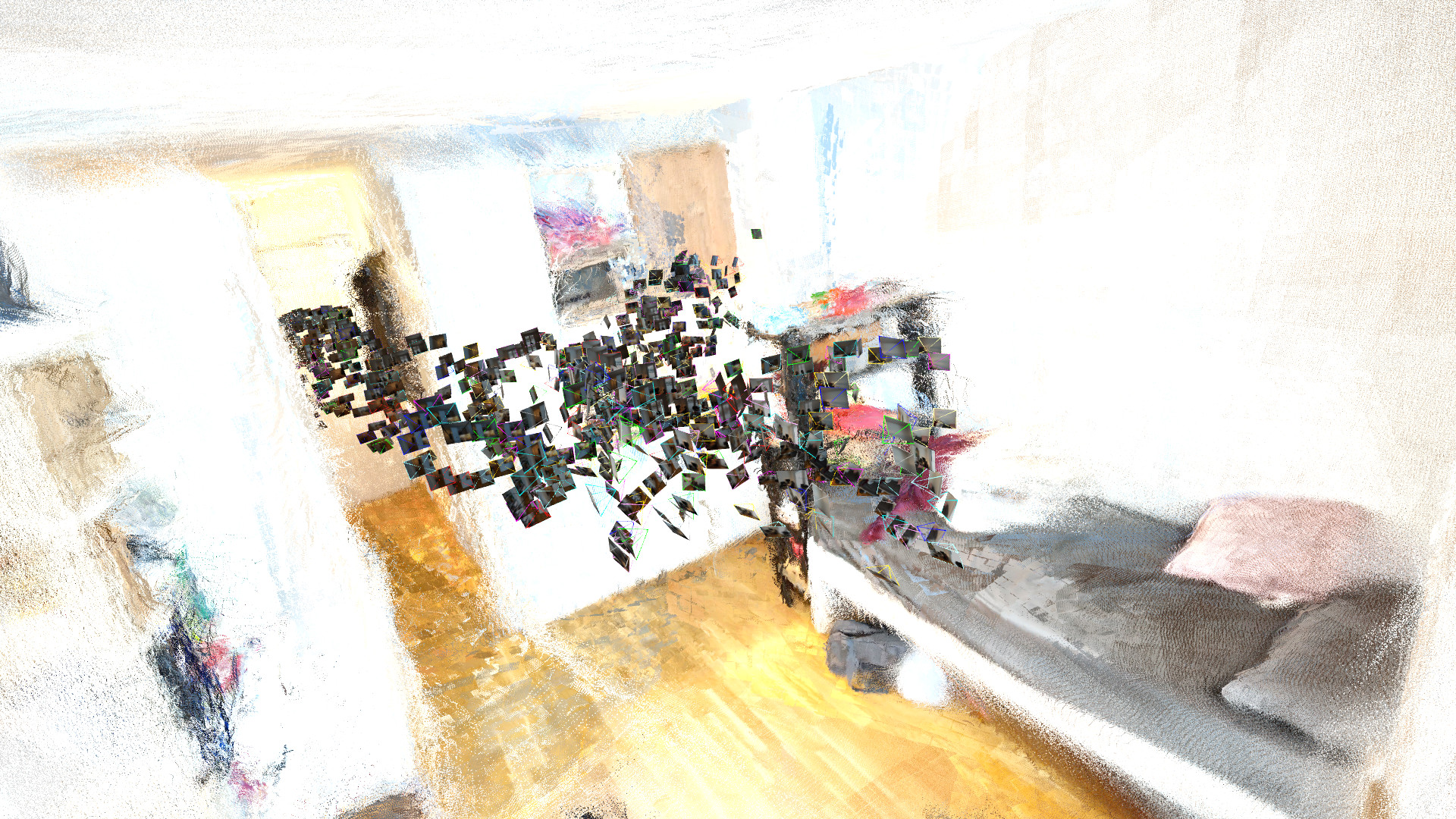} \includegraphics[width=0.32\linewidth, trim=0 0 0 0, clip]{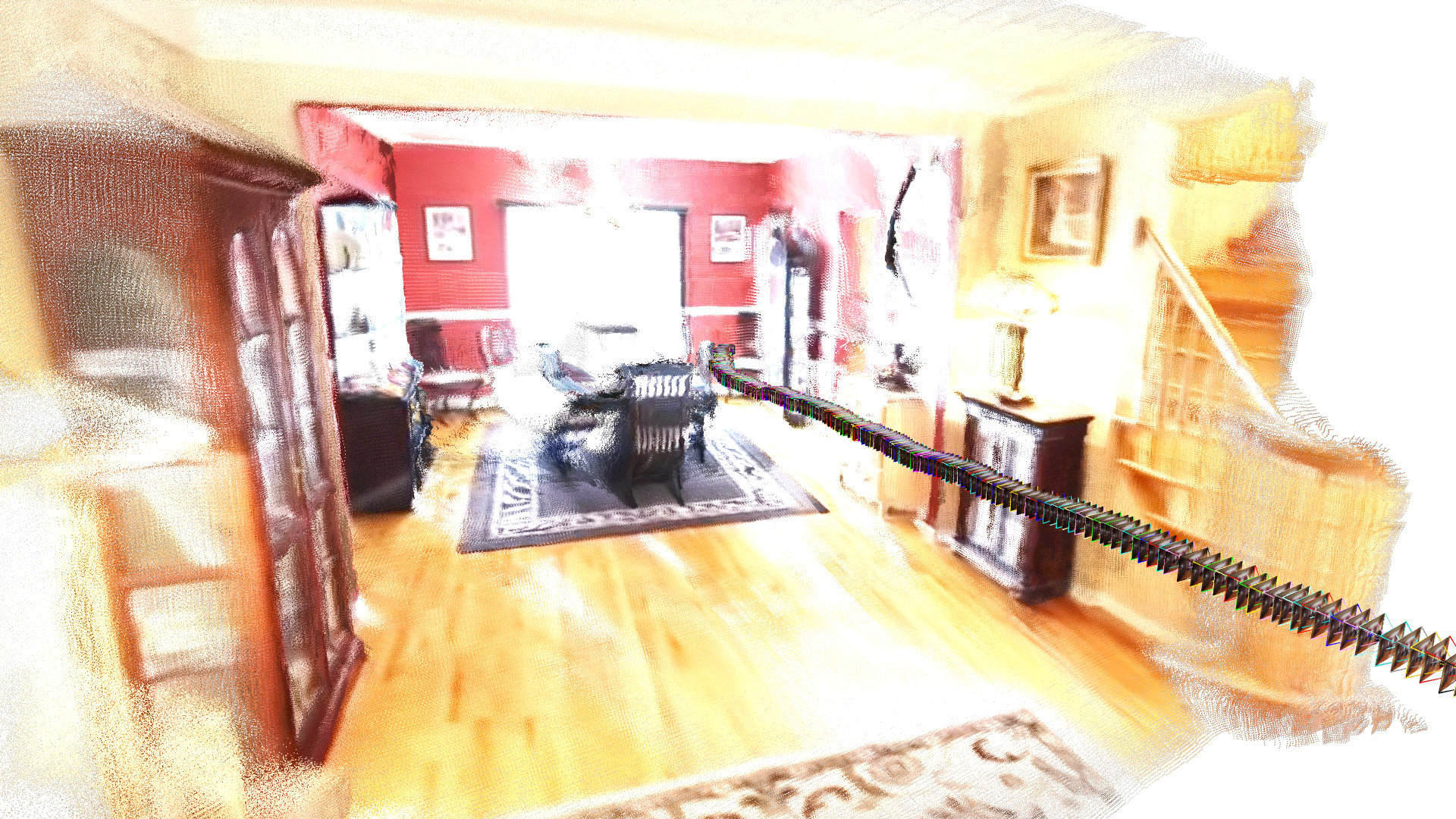} \includegraphics[width=0.32\linewidth, trim=0 0 0 0, clip]{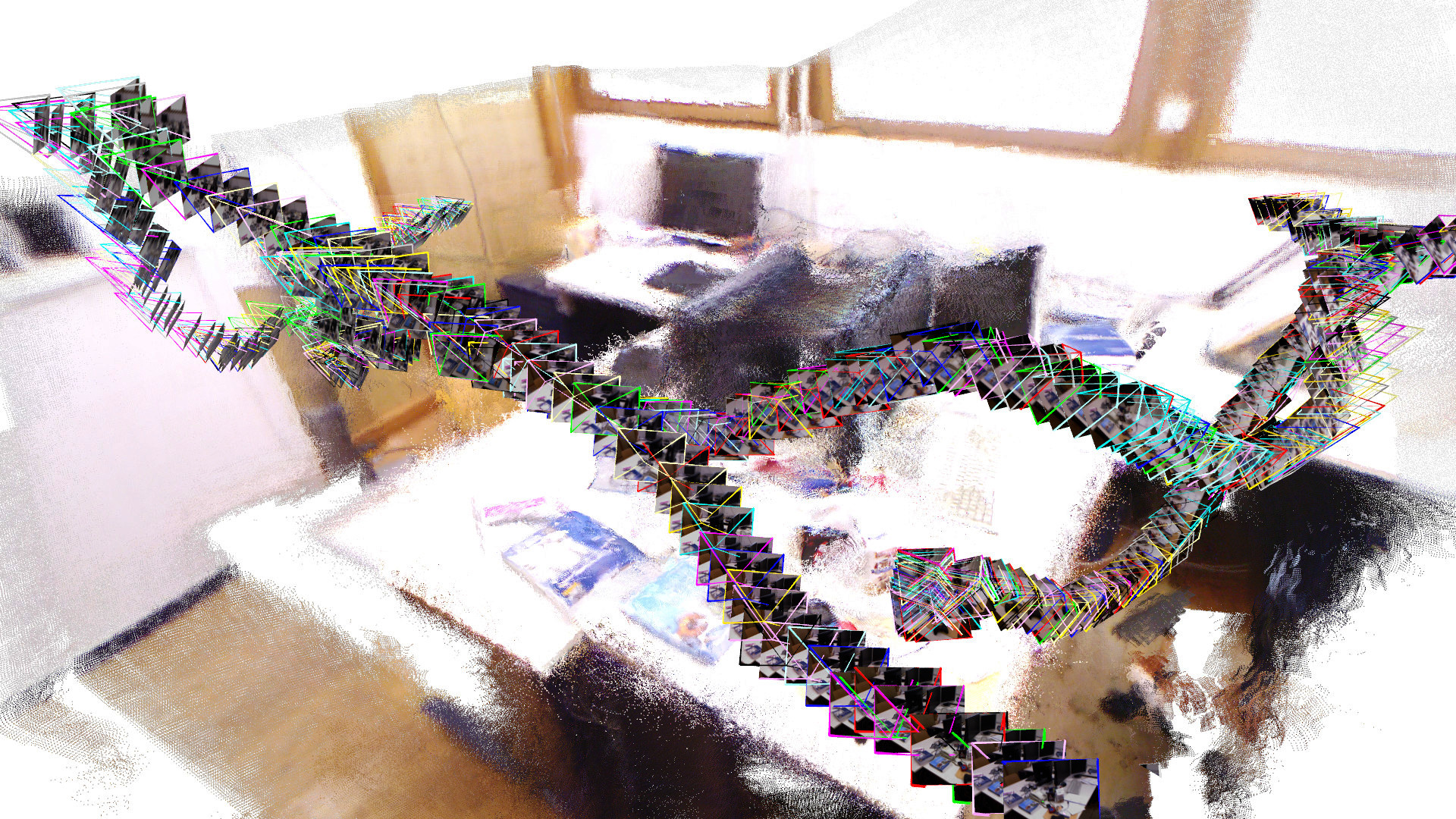}  \includegraphics[width=0.32\linewidth, trim=0 0 0 0, clip]{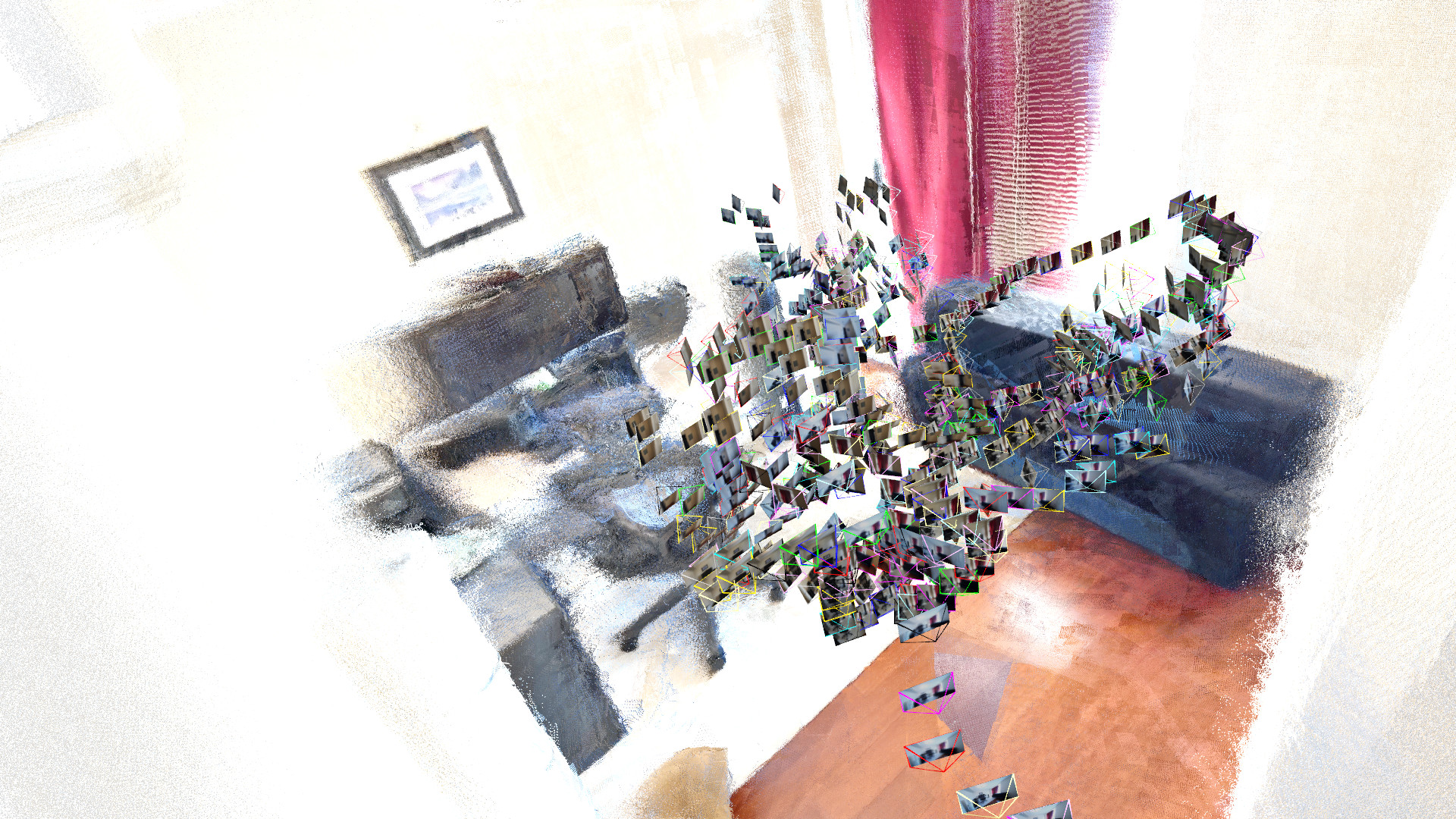}    \includegraphics[width=0.32\linewidth, trim=0 0 0 0, clip]{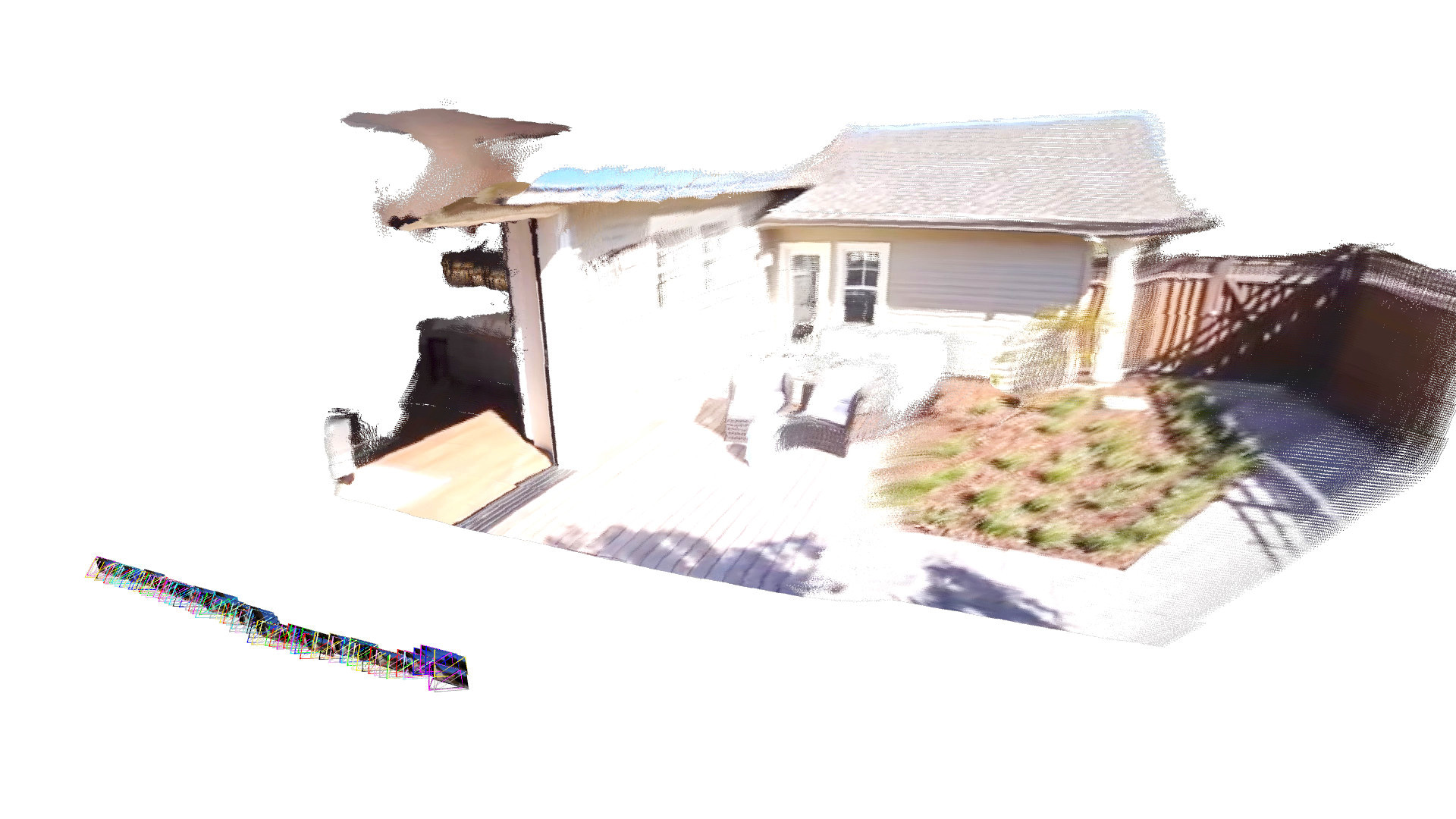} \includegraphics[width=0.32\linewidth, trim=0 0 0 0, clip]{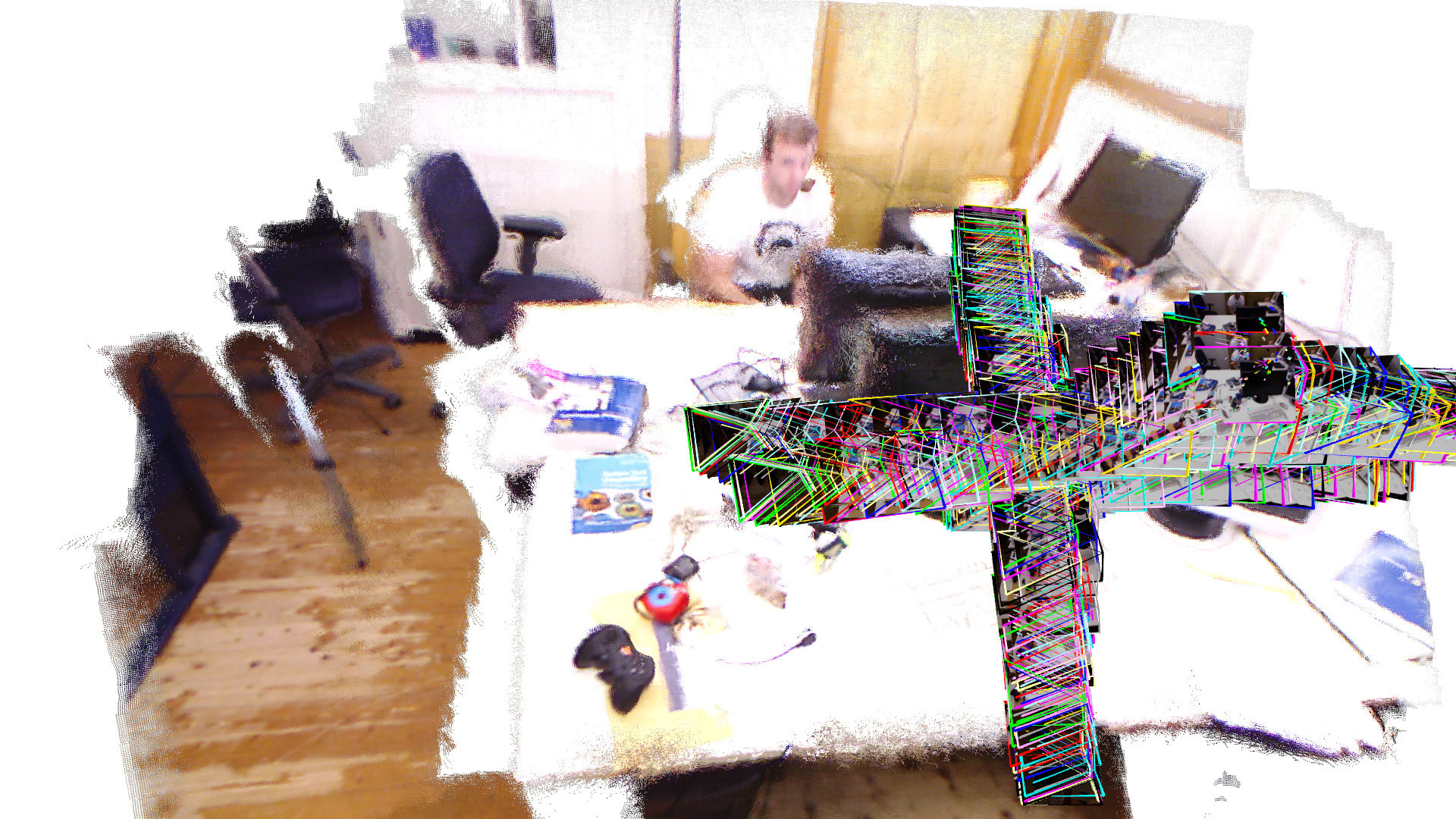}
    \caption{
        Qualitative example of \munster{} reconstructions of ScanNet++ \cite{scannet++}  \textbf{0d2ee665b} and \textbf{a24f64f7fb} scenes (\textit{left column}) with varying focal lengths. RealEstate10K \cite{ZhouTOG18Realestate10K} \textbf{000c3ab189999a83} and \textbf{00ca5123d8ff6f83}
        scenes (\textit{center column}). TUM RGB-D \cite{Sturm2012ASystems} freiburg1  \textbf{desk} and \textbf{xyz} scenes (\textit{right column}).
    }
    \label{fig:qualitative1_supmat}
\end{figure*}

\begin{figure*}
    \centering
     \includegraphics[width=0.48\linewidth, trim=0 0 0 0, clip]{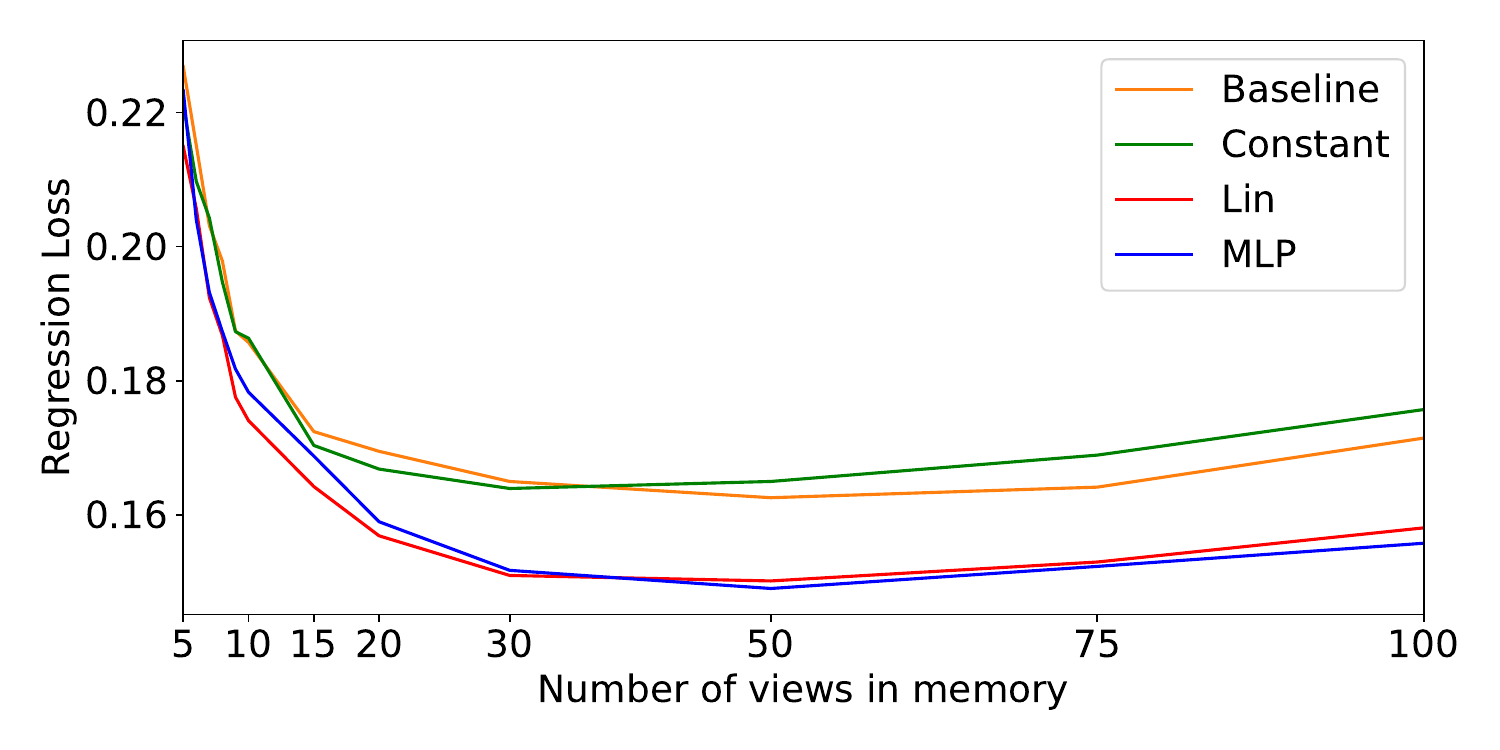}
   \includegraphics[width=0.48\linewidth, trim=0 0 0 0, clip]{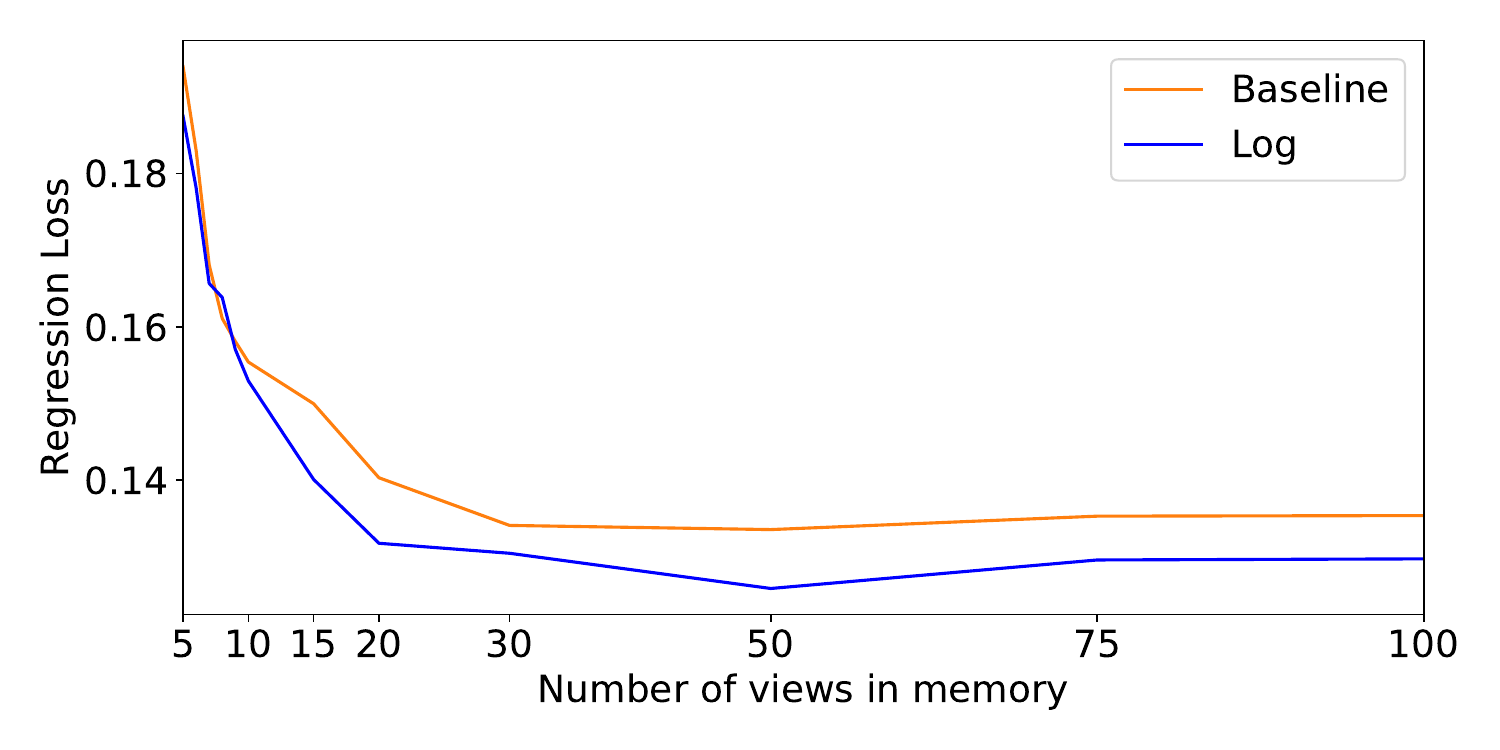}
    \caption{ \textbf{Median absolute regression error} for ScanNet scenes for $100$ views rendered with a varying number of views in memory.
        \textit{(left)} Different injection $\awa$ architecture choices.
        \textit{(right)} Comparison between a log and the default \master{} loss for metric pointmap regression.  
    }
    \label{fig:ablation}
\end{figure*}

\mypar{Results.} We report the results of these ablations in Tab.~\ref{tab:ablation_sym}. First, we show that RoPE in the CA block does not matter for performance. We can thus remove it without loss in prediction quality.
Then, when using a symmetric decoder without learnable $B$ encoding collapses in \red{red}. It is rather natural as the network needs to distinguish between the views, justifying the need for the learnable embedding $B$ that achieves a performance similar to that of \duster.
Overall, it is clear that a symmetric architecture does not impact the performance despite halving the amount of trainable parameters in the decoder.

\begin{table}
\begin{center}
\renewcommand\arraystretch{1.2}
\setlength{\tabcolsep}{1pt} 
\tiny
\resizebox{0.7\columnwidth}{!}{
\begin{tabular}{l|c}
\toprule 
 & Regression Loss $\downarrow$  \\ 
\midrule

\textbf{Baseline} & 0.193 \\ 
\textbf{No RoPE in CA} & 0.191 \\ 
\textbf{Symmetric(none)} & \textcolor{red}{0.560} \\ 
\textbf{Symmetric(embed)} & 0.192 \\ 

\bottomrule

\end{tabular}}
\caption{Analysis of the effects of our proposed simplifications on \duster{} in the pairwise setting. }
\label{tab:ablation_sym}
\end{center}
\end{table}

\subsection{\munster{} architecture and loss}
\label{sec:archi}
We verified in Appendix~\ref{supssec:symduster} that a symmetric architecture does not impact the performance in the binocular case. How to handle $N$ views predictions still remains a challenge however, as directly processing all the views would involve an intractable number of tokens for both training and testing. For this reason, we favor a memory-based approach where only a subset of image tokens is used in memory. Images are thus processed sequentially and we keep the computational complexity low. The following contains the ablations that lead our choices in the design of~\munster{} training in the multi-view setting. In more details, we demonstrate the importance of the global 3D feedback from the terminal layer to earlier layers. We also analyze the effect of computing the loss in log space, that becomes crucial for larger scenes.

\mypar{Setup and metric.}  
For the $N$ views case, we choose to train all variants on $10$-tuples and validate on $100$-tuples from the ScanNet++~\cite{scannet++} dataset, split into a training set of $2$M $10$-tuples and a validation set of $300$ $100$-tuples, totaling for the latter $30$K images from $50$ different scenes. We report the metric 3D regression loss with pointmaps expressed in the coordinate frame of the first view, meaning no alignment between the prediction and the \emph{ground-truth} is required. All results are \emph{renderings} $N {=} 100$ images with a varying number of images stored in memory ($5 {\leq} n {\leq} N$). We plot in Fig.\ref{fig:ablation}
the median scene regression error over the 
validation subset.

\begin{table*}
\begin{center}
\renewcommand\arraystretch{1.2}
\setlength{\tabcolsep}{1pt} 
\tiny
\hspace{-3mm}
\resizebox{\textwidth}{!}{
\begin{tabular}{lcc|rlrlrlc|rlrlrlc|rlrlrlc|c|c}
\specialrule{1pt}{0.5pt}{0.5pt}
{\multirow{3}{*}{Methods}} &&  &  \multicolumn{6}{c}{7-Scenes} & &\multicolumn{6}{c}{NRGBD}
 & &\multicolumn{6}{c}{DTU} & & \\
&   &&
\multicolumn{2}{c}{Acc $\downarrow$} 
&   \multicolumn{2}{c}{Comp $\downarrow$} 
&   \multicolumn{2}{c}{NC $\uparrow$}
&  & \multicolumn{2}{c}{Acc $\downarrow$} &
  \multicolumn{2}{c}{Comp $\downarrow$} & 
  \multicolumn{2}{c}{NC $\uparrow$} 
   & & \multicolumn{2}{c}{Acc $\downarrow$} &
  \multicolumn{2}{c}{Comp $\downarrow$} & 
  \multicolumn{2}{c}{NC $\uparrow$} 
& & FPS  & Mem \\
 & $n$ & $s$ &  Mean & Med. & Mean & Med. & Mean & Med. & & Mean & Med. & Mean & Med. & Mean & Med.
  & & Mean & Med. & Mean & Med. & Mean & Med. & & &\\
\specialrule{1pt}{0.5pt}{0.5pt}

F-Recon \cite{xu2023frozenrecon}  &&& 0.124 &  0.076 &  0.055 & 0.023 & 0.619 & 0.688 & &0.285 & 0.206 & 0.151 &  0.063 &  0.655 &  0.758 && - & - & - & - & - & - & &($\leq 1$) & \\
\duster{}-224 \cite{duster} &&& 0.029 & 0.012 &  0.028 & 0.009 & \textbf{0.668} & \textbf{0.768} 
& & 0.054 & 0.025 & 0.032 & 0.010 & \textbf{0.802} &  \textbf{0.953} 
& & \textbf{2.296} & \textbf{1.297} & 2.158 & 1.002 & \textbf{0.747} & \textbf{0.848}
&  & 0.74 (0.78) & 38.1G\\
\spanner{} \cite{WangX24Spanner3DReconstructionWithSpatialMemory} &&& 0.034 &  0.015 &  \textbf{0.024} & 0.009 & 0.664 & 0.763 & & 0.069 & 0.032 & 0.029 &  0.011 & 0.778 &  0.937 & 
& 4.785 & 2.268 & 2.743 & 1.295 &  0.721 & 0.823
& &  27.38 (65.49) & 5.0G\\
\specialrule{0.5pt}{0.5pt}{0.5pt}

\munster{}-224 & 10 & 1 & 0.037 & 0.017 & 0.026 & 0.010 & 0.654 & 0.741 & & 0.064 & 0.028 & 0.032 & 0.011 & 0.778 & 0.922 & & 3.256 & 1.863 & 2.193 & 0.995 & 0.715 & 0.815 & & \textbf{86.31} & \textbf{2.5G} \\

\munster{}-224 & 20 & 1  & 0.030 & 0.013 & 0.026 & 0.010 & 0.662 & 0.753 & & 0.061 & 0.026 & 0.029 & 0.011 & 0.786 & 0.928 & & 3.256 & 1.863 & 2.193 & 0.995 & 0.715 & 0.815 & & 59.51 & 2.9G \\
\munster{}-224 & all & 1 & 0.028 & 0.012 & 0.027 & 0.010 & 0.665 & 0.758 & & 0.062 & 0.025 & 0.031 & 0.012 & 0.788 & 0.930 & & 3.256 & 1.863 & 2.193 & 0.995 & 0.715 & 0.815 & & 40.41 & 4.1G \\
\munster{}-224 & all & 5  & 0.035 & 0.015 & 0.027 & 0.011 & 0.662 & 0.752 & & 0.059 & 0.025 & 0.029 & 0.011 & 0.792 & 0.934 & & 3.261 & 1.855 & 2.157 & 1.016 & 0.715 & 0.815 & & 68.83 & 4.1G \\

\munster{}-224 & all & 10 & 0.037 & 0.016 & 0.028 & 0.011 & 0.661 & 0.751 & & 0.057 & 0.025 & 0.028 & 0.011 & 0.792 & 0.934 & & 3.269 & 1.805 & 2.147 & 0.978 & 0.715 & 0.816 & & 72.59 & 4.8G \\

\specialrule{1pt}{0.5pt}{0.5pt}
\munster{}-512 & 10 & 1 & 0.028 & \textbf{0.009} & 0.025 & \textbf{0.007} & 0.617 & 0.682 & & 0.052 & 0.023 & \textbf{0.019} & 0.008 & 0.764 & 0.907 & & 3.261 & 1.681 & \textbf{1.965} & \textbf{0.765} & 0.661 & 0.741 & & 25.01 &  4.3G \\
\munster{}-512 & 20 & 1 & \textbf{0.025} & \textbf{0.009} & 0.026 & 0.008 & 0.617 & 0.683 & & 0.048 & 0.022 & \textbf{0.019} & 0.008 & 0.769 & 0.911 & & 3.261 & 1.681 & \textbf{1.965} & \textbf{0.765} & 0.661 & 0.741 & & 17.37 & 5.3G \\
\munster{}-512 & all & 1 & 0.026 & \textbf{0.009} &  0.027 & 0.009 & 0.617 & 0.682 & & 0.048 & 0.022 & 0.020 & 0.008 & 0.768 &  0.911 & & 3.261 & 1.681 & \textbf{1.965} & \textbf{0.765} & 0.661 & 0.741 & & 12.10 & 8.1G \\
\munster{}-512 & all & 5 &  0.036 & 0.014 & 0.025 & 0.008 & 0.616 & 0.680 & & 0.048 & \textbf{0.021} & \textbf{0.019} & 0.008 & 0.770 & 0.912 & & 3.353 & 1.741 & 1.982 & 0.772 & 0.663 & 0.742 & & 13.52 &  9.9G \\
\munster{}-512 & all & 10 & 0.042 & 0.016 & \textbf{0.024} & 0.008 & 0.615  & 0.679 & & \textbf{0.047} & \textbf{0.021} & \textbf{0.019} & \textbf{0.007} & 0.769 & 0.910 & & 3.493 & 1.888 & 2.046 & 0.809 & 0.664 & 0.745 & & 12.79 & 10.5G \\

\specialrule{1pt}{0.5pt}{0.5pt}
\end{tabular}}

\normalsize
\caption{
\textbf{Comparison with \spanner{}}. We evaluate \munster{} for different maximum size of memory ($n$) and different number of images at once when updating the memory ($s$). For $s{=}1$, we initialize with two images to match the training configuration. FPS numbers in parenthesis are from \cite{WangX24Spanner3DReconstructionWithSpatialMemory}. For \duster{}, our FPS and GPU memory numbers were obtained with the 224 linear model and a complete graph.}
\label{tab:spannercompar_suppmat}
\end{center}
\end{table*}

\mypar{3D feedback.}
We ablate in Fig.~\ref{fig:ablation} (\textit{left}) different variants of Global 3D Feedback from section \ref{sec:causality}. In particular, we demonstrate that the \textbf{Baseline} architecture without feedback is far inferior to the \textbf{MLP} injection from the tokens of the last memory block presented in the main paper (Sec.~\ref{sec:causality} and Fig.~\ref{fig:munst3r_offset}). As a reminder, this approach injects the terminal memory state $L{-}1$ to all others with a LayerNorm followed by a $2$ layer MLP, with a fourfold increase in the hidden dimensions. Interestingly, \textbf{Lin} injection with a simple linear layer instead of an MLP performs almost as well except when scaling the memory size. Since we expect training on all datasets to be a harder task, we favor a slightly more elaborate injection mechanism. We also verify that a simple learnable parameter \textbf{Constant} is not sufficient and actually degrades the performance, demonstrating that our feedback layer is effectively fetching information from the terminal memory layer and not simply adding a constant bias to the memory tokens. 
Importantly, a natural choice would be to inject global information from the final layer $L$ of the network, \ie the one right before the prediction head. We find that this method leads to a significantly degraded performance. We hypothesize that using $L$ might be too constrained by $\head{}$, and therefore it is better to add feedback from layer $L{-}1$, the terminal memory layer.

\mypar{Computing the loss in log space.}
To analyze the choice of log space loss in section \ref{sec:details}, we compare two \munster{} models trained following the recipe of section \ref{sec:details}.

The \textbf{Log} model is trained with the log space loss (Eq.~\ref{eq:log_reg})  while \textbf{Baseline} is trained with the default metric regression loss from \master{}. Again, we show the results for varying  number of images used in the memory
in Fig.~\ref{fig:ablation} (Right). We can notice a clear gap between the two losses when going above $10$ views in memory, the model being trained with the loss in log space providing a significantly better scaling to more views. For instance with $n\seq 50$ views in memory the baseline achieves $13.4$cm accuracy \vs $12.6$ for the log space loss. An interpretation is that the log space loss allows the network to regress larger values which is crucial for more input views as they will most often observe larger parts of the scenes.

\subsection{More views at inference}
\label{sec:runtime}
Beyond architectural choices, an important result of the previous section is the scalability of \munster{} to more views than ever seen during training, as explained in section \ref{sec:munst3r_base}. 
In particular, we can use our architecture to robustly predict from $n {\gg} 10 $ views in memory despite being trained only with $n {\leq} 10$ as seen in Fig.~\ref{fig:ablation}. Interestingly, it can also process $s {\gg} 1$ views at the same time while having been trained to predict $s{=}1$ views at the same time.

In this section, we study the impact of the number of views in the memory  ($n$) and the number of views used at once ($s$) when updating the memory
in \munster{}. 
These comparisons can be seen on the one hand in 
Tab.~\ref{tab:spannercompar_suppmat} where we complement
Tab.~\ref{tab:spannercompar} with experiments on \munster{}
varying $n$ and $s$ during inference 
(see columns 2 and 3). 

\begin{table*}
\begin{center}
\renewcommand\arraystretch{1.2}
\setlength{\tabcolsep}{1pt} 
\tiny
\hspace{-3mm}
\resizebox{\textwidth}{!}{
\begin{tabular}{cl|cccccccccccccccccccccccccccccccccc|c}
\specialrule{1pt}{0.5pt}{0.5pt}
    
& & \multicolumn{1}{|c|}{fr1} & \multicolumn{14}{|c|}{fr2} & \multicolumn{19}{|c|}{fr3} & \\
& &  \multicolumn{1}{|c|}{floor} &
360\_hsph & 360\_kid & coke & desk\_p & dishes & flow\_bqt & flow\_br & met\_sph & met\_sph2 & pion\_360 & pion\_slm & pion\_slm2 & pion\_slm3 &  \multicolumn{1}{c|}{rpy} & 
cab & lcab & ns\_nt\_far & ns\_nt\_loop & ns\_tex\_far & ns\_tex\_loop & sit\_hsph & sit\_rpy & sit\_stat & sit\_xyz & str\_nt\_far & str\_nt\_near & str\_tex\_far & str\_tex\_near & teddy & walk\_hsph & walk\_rpy & walk\_stat & walk\_xyz & \textbf{Avg} \\
\specialrule{1pt}{0.5pt}{0.5pt}
\multicolumn{37}{c}{RMSE ATE Tracking Error [cm] $\downarrow$} \\
\specialrule{1pt}{0.5pt}{0.5pt}

D  & GlORIE-VO * \cite{ZhangX24GlORIESLAM} & 15.5 & 84.2 & 108.6 & \textbf{5.4} & \textbf{1.9} & \textbf{3.0} & \textbf{2.7} & 3.5 & \textbf{5.6} & \textbf{6.0} & \textbf{31.5} & 87.1 & 130.1 & 170.8 & \textbf{0.4} & 2.0 & 5.8 & \textbf{2.5} & \textbf{73.6} & \textbf{2.2} & \textbf{4.0} & \textbf{3.4} & \textbf{1.8} & - & \textbf{0.9} & \textbf{1.6} & \textbf{1.9} & \textbf{1.0} & \textbf{1.9} & 3.7 & \textbf{1.9} & \textbf{5.5} & 2.3 & \textbf{1.5} & 23.4 \\
\hline
& \spanner{} \cite{WangX24Spanner3DReconstructionWithSpatialMemory} & 75.6 & 116.9 & 144.5 & 64.7 & 12.7 & 9.8 & 19.5 & 28.9 & 66.2 & 76.6 & 154.7 & 193.9 & 200.5 & 206.5 & 3.9 & 3.3 & 154.6 & 56.4 & 138.3 & 23.4 & 191.9 & 26.0 & 6.6 & 2.8 & 7.0 & 27.3 & 76.7 & 52.1 & 76.6 & 65.5 & 33.3 & 15.0 & 2.2 & 9.2 & 68.9 \\
 
U  & \munster{}-224-C & 8.8 & 88.4 & 130.8 & 11.5 & 5.0 & 7.2 & 7.9 & 3.2 & 13.2 & 19.0 & 71.1 & 47.7 & 140.3 & 45.4 & 2.0 & 2.2 & 5.7 & 12.0 & 142.2 & 12.7 & 11.8 & 32.9 & 6.5 & 2.7 & 5.6 & 4.4 & 15.7 & 5.3 & 7.0 & 5.8 & 37.2 & 18.0 & 1.7 & 4.7 & 27.5 \\
   & \munster{}-224 & \textbf{5.0} & \textbf{79.6} & \textbf{74.1} & 8.6 & 3.0 & 5.5 & 5.0 & \textbf{2.2} & 11.4 & 10.9 & 40.5 & \textbf{18.1} & \textbf{69.1} & \textbf{41.8} & 1.4 & \textbf{1.6} & \textbf{4.6} & 8.6 & 117.2 & 10.8 & 10.6 & 30.9 & 6.5 & \textbf{2.0} & 4.2 & 3.2 & 10.2 & 4.1 & 6.4 & \textbf{3.2} & 28.3 & 17.8 & \textbf{1.4} & 3.1 & \textbf{19.1}  \\
\specialrule{1pt}{0.5pt}{0.5pt}
\multicolumn{37}{c}{Vertical FoV Error (in degrees) $\downarrow$} \\
\specialrule{1pt}{0.5pt}{0.5pt}
   & \spanner{} \cite{WangX24Spanner3DReconstructionWithSpatialMemory} & 31.46 & 8.85 & 2.98 & 11.94 & 11.78 & 11.97 & 13.96 & 16.54 & 9.15 & 10.65 & 10.86 & 11.8 & 11.64 & 0.86 & 12.41 & 9.53 & 9.53 & 7.55 & 8.7 & 10.54 & 17.37 & 9.6 & 9.78 & 9.77 & 8.83 & 12.16 & 17.86 & 10.57 & 9.87 & 10.78 & 9.41 & 9.16 & 9.54 & 9.46 & 11.08 \\
U  & \munster{}-224-C & 0.15 & \textbf{0.30} & \textbf{0.95} & \textbf{0.90} & \textbf{0.44} & \textbf{0.99} & \textbf{0.04} & \textbf{1.23} & \textbf{0.52} & \textbf{0.40} & 0.69 & \textbf{0.66} & \textbf{1.26} & \textbf{0.60} & \textbf{0.45} & \textbf{0.61} & \textbf{0.75} & \textbf{2.25} & \textbf{3.4} & \textbf{5.47} & \textbf{5.09} & \textbf{0.43} & 0.49 & \textbf{2.86} & \textbf{3.16} & \textbf{0.70} & \textbf{0.42} & \textbf{0.64} & \textbf{0.74} & \textbf{0.09} & \textbf{0.07} & \textbf{0.80} & \textbf{2.08} & \textbf{1.72} & \textbf{1.22} \\
  & \munster{}-224 & \textbf{0.14} & \textbf{0.30} & \textbf{0.95} & \textbf{0.90} & \textbf{0.44} & \textbf{0.99} & \textbf{0.04} & \textbf{1.23} & \textbf{0.52} & \textbf{0.40} & \textbf{0.68} & 0.70 & \textbf{1.26} & 0.61 & \textbf{0.45} & \textbf{0.61} & \textbf{0.75} & \textbf{2.25} & \textbf{3.4} & \textbf{5.47} & \textbf{5.09} & \textbf{0.43} & \textbf{0.47} & \textbf{2.86} & \textbf{3.16} & \textbf{0.70} & \textbf{0.42} & \textbf{0.64} & \textbf{0.74} & \textbf{0.09} & \textbf{0.07} & \textbf{0.80} & \textbf{2.08} & \textbf{1.72} & \textbf{1.22} \\
   
\specialrule{1pt}{0.5pt}{0.5pt}
\multicolumn{37}{c}{Scale Error $\downarrow$} \\
\specialrule{1pt}{0.5pt}{0.5pt}
D  & GlORIE-VO * \cite{ZhangX24GlORIESLAM} & 0.85 & 19.14 & 3.27 & 0.62 & 1.27 & 0.88 & 0.72 & 1.61 & 0.8 & 1.82 & 3.94 & 1.11 & 0.28 & 0.86 & 1.62 & 1.51 & 3.19 & 1.96 & 0.4 & 2.56 & 1.4 & 2.39 & 2.25 & 0.63 & 2.38 & 2.25 & 1.04 & 2.53 & 1.34 & 1.08 & 2.79 & 3.41 & 0.15 & 12.8 &  2.50 \\
\hline
& \spanner{} \cite{WangX24Spanner3DReconstructionWithSpatialMemory}& 0.9 & 3.2 & 4.31 & 2.8 & 5.25 & 3.54 & 3.57 & 2.91 & 3.36 & 1.81 & 3.96 & 5.92 & 2.84 & 4.24 & 2.66 & 4.36 & 10.25 & 9.0 & 7.41 & 9.59 & 2.3 & 3.28 & 0.2 & 4.33 & 7.43 & 5.59 & 5.12 & 6.83 & 7.04 & 1.82 & 3.54 & 0.81 & 1.59 & 6.82 & 4.37 \\
U  & \munster{}-224-C & \textbf{0.19} & 0.19 & 0.36 & 0.07 & 0.09 & 0.04 & 0.02 & 0.19 & \textbf{0.01} & 0.63 & 0.46 & 0.12 & 0.2 & 0.14 & 0.11 & \textbf{0.05} & \textbf{0.36} & \textbf{0.04} & 0.52 & 1.21 & 0.52 & 0.68 & 0.89 & 0.28 & \textbf{0.01} & \textbf{0.03} & \textbf{0.54} & 0.49 & \textbf{0.13} & \textbf{0.28} & 0.49 & 0.94 & 0.62 & \textbf{0.02} & 0.32 \\
& \munster{}-224 & 0.22 & \textbf{0.1} & \textbf{0.07} & \textbf{0.00} & \textbf{0.07} & \textbf{0.02} & \textbf{0.00} & \textbf{0.18} & 0.02 & \textbf{0.26} & \textbf{0.36} & \textbf{0.07} & \textbf{0.09} & \textbf{0.09} & \textbf{0.00} & 0.06 & 0.38 & \textbf{0.04} & \textbf{0.42} & \textbf{1.18} & \textbf{0.51} & \textbf{0.59} & \textbf{0.88} & \textbf{0.04} & 0.02 & 0.04 & \textbf{0.54} & \textbf{0.46} & 0.15 & \textbf{0.28} & \textbf{0.24} & \textbf{0.89} & \textbf{0.47} & \textbf{0.02} & \textbf{0.26} \\
\specialrule{1pt}{0.5pt}{0.5pt}
\end{tabular}
}
\normalsize
\caption{\textbf{Detailed results on TUM RGBD~\cite{Sturm2012ASystems}} (34 sequences not included in Tab.~1, and summarized in Tab.~2 of the main text). One Dense (D) versus dense unconstrained (U) methods on TUM-RGBD SLAM benchmark. (*) GlORIE-SLAM \cite{ZhangX24GlORIESLAM} was re-run without Loop Closure and global Bundle Adjustment.
}
\label{tab:tum_full}
\end{center}
\vspace*{-1mm}
\end{table*}

\begin{table*}[ht]
\begin{center}
\renewcommand\arraystretch{1.2}
\setlength{\tabcolsep}{1pt} 
\tiny
\resizebox{1.0\textwidth}{!}{ 
\begin{tabular}{l|ccccccccccccccccccccccccccccccc|c}
\specialrule{1pt}{0.5pt}{0.5pt}
&cash1 & cash2 & cei1 & de3 & dech1 & eiflli & eillc1 & eillc2 & eiglc3 & mq1 & mq4 & mqfa1 & mqfa2 & mqhe & mo1 & plr2 & plt3 & plt4 & plt5 & pltsc1 & ref1 & rep & sfmbe & sfmgr & sfmhl & sfmlr1 & sf2 & sf3 & sfsh & vili1 & vili2 & Avg \\
\specialrule{1pt}{0.5pt}{0.5pt}
\multicolumn{33}{c}{RMSE ATE Tracking Error [m] $\downarrow$} \\
\specialrule{1pt}{0.5pt}{0.5pt}
\spanner{}\cite{WangX24Spanner3DReconstructionWithSpatialMemory} &0.05 & 0.08 & 2.3 & 0.96 & 1.25 & 1.41 & 0.68 & 0.12 & 0.76 & 0.73 & 1.09 & 0.5 & \textbf{0.03} & 0.25 & 0.65 & 0.38 & 0.18 & 0.05 & 0.25 & 0.95 & \textbf{0.39} & 0.91 & 0.5 & 5.93 & 7.72 & 1.36 & 0.1 & 0.14 & 0.1 & 0.84 & 0.18 & 0.99 \\
\munster{}-224-C&0.06 & 0.05 & 2.01 & \textbf{0.89} & \textbf{0.81} & 1.32 & 0.73 & 0.09 & 0.93 & 0.79 & 0.6 & \textbf{0.36} & 0.07 & 0.19 & \textbf{0.18} & \textbf{0.07} & 0.28 & \textbf{0.03} & 0.24 & \textbf{0.08} & \textbf{0.39} & 0.76 & 0.14 & \textbf{4.32} & 6.12 & \textbf{0.72} & 0.36 & 0.35 & 0.1 & 0.63 & 0.26 & 0.77\\
\munster{}-224&\textbf{0.04} & \textbf{0.03} & \textbf{1.98} & 1.02 & 0.85 & \textbf{1.29} & \textbf{0.59} & \textbf{0.08} & \textbf{0.64} & \textbf{0.32} & \textbf{0.33} & 0.42 & 0.04 & \textbf{0.16} & 0.26 & \textbf{0.07} & \textbf{0.12} & 0.06 & \textbf{0.2} & 0.1 & \textbf{0.39} & \textbf{0.31} & \textbf{0.11} & 5.96 & \textbf{5.0} & \textbf{0.72} & \textbf{0.05} & \textbf{0.08} & \textbf{0.09} & \textbf{0.28} & \textbf{0.14} & \textbf{0.70} \\ 
\specialrule{1pt}{0.5pt}{0.5pt}
\multicolumn{33}{c}{Vertical FoV Error (in degrees)$\downarrow$} \\
\specialrule{1pt}{0.5pt}{0.5pt}
\spanner{} \cite{WangX24Spanner3DReconstructionWithSpatialMemory}&21.26 & 17.24 & 27.83 & 26.53 & 26.16 & 28.28 & 25.55 & 26.28 & 25.78 & 28.63 & 30.33 & 67.47 & 26.86 & 27.41 & 25.92 & 28.64 & 26.01 & 27.0 & 25.96 & 26.5 & 25.51 & 25.61 & 16.62 & 26.09 & 26.94 & 26.61 & 25.51 & 25.37 & 25.62 & 67.27 & 26.95 & 28.50\\
\munster{}-224-C&3.68 & 1.04 & 5.89 & 2.01 & \textbf{1.35} & 13.67 & 2.56 & 1.82 & 1.93 & 5.18 & 6.28 & \textbf{9.25} & \textbf{5.51} & 7.79 & 1.75 & 10.39 & 14.07 & 11.47 & 12.37 & 1.12 & 3.56 & \textbf{0.03} & 2.04 & 4.84 & 4.28 & 1.42 & 1.65 & 2.8 & \textbf{2.43} & 11.52 & 13.92 & 5.40 \\
\munster{}-224&\textbf{0.63} & \textbf{1.01} & \textbf{1.4} & \textbf{1.0} & 1.39 & \textbf{10.97} & \textbf{1.17} & \textbf{1.16} & \textbf{1.07} & \textbf{3.27} & \textbf{2.12} & 12.15 & 6.84 & \textbf{3.76} & \textbf{0.81} & \textbf{7.46} & \textbf{4.28} & \textbf{8.04} & \textbf{6.56} & \textbf{1.0} & \textbf{3.1} & 1.6 & \textbf{0.33} & \textbf{0.72} & \textbf{2.63} & \textbf{1.25} & \textbf{0.94} & \textbf{0.11} & 2.66 & \textbf{4.21} & \textbf{4.58} & \textbf{3.17}\\
\specialrule{1pt}{0.5pt}{0.5pt}
\multicolumn{33}{c}{Scale Error (ratio of GT scale) $\downarrow$} \\ 
\specialrule{1pt}{0.5pt}{0.5pt}
\spanner{} \cite{WangX24Spanner3DReconstructionWithSpatialMemory}&1.03 & 0.79 & 1.17 & \textbf{0.4} & 1.51 & 2.33 & 4.87 & 5.44 & 4.02 & 1.8 & 2.9 & 0.64 & 1.67 & 0.86 & 4.22 & \textbf{0.25} & 1.14 & 0.16 & \textbf{0.05} & 3.31 & 0.63 & 0.3 & 3.42 & 8.06 & 13.27 & 7.2 & 3.62 & 4.66 & 1.85 & 3.05 & 2.11 & 2.79 \\
\munster{}-224-C&0.49 & \textbf{0.32} & \textbf{0.28} & 0.59 & \textbf{0.32} & 0.67 & 0.25 & 0.19 & 0.27 & \textbf{0.48} & \textbf{0.26} & 0.63 & \textbf{0.53} & \textbf{0.59} & 0.13 & 0.65 & \textbf{0.01} & \textbf{0.62} & 0.54 & 0.29 & \textbf{0.57} & 0.4 & \textbf{0.19} & 0.49 & \textbf{0.04} & 0.02 & 0.13 & \textbf{0.06} & 0.3 & \textbf{0.21} & 0.65 & 0.36\\
\munster{}-224&\textbf{0.32} & 0.4 & 0.49 & 0.77 & 0.41 & \textbf{0.61} & \textbf{0.05} & \textbf{0.12} & \textbf{0.04} & 0.67 & 0.41 & \textbf{0.01} & 0.74 & 0.72 & \textbf{0.05} & 0.48 & 0.43 & 0.82 & 0.62 & \textbf{0.11} & 0.67 & \textbf{0.05} & \textbf{0.19} & \textbf{0.09} & 0.37 & \textbf{0.01} & \textbf{0.11} & \textbf{0.06} & \textbf{0.21} & 0.34 & \textbf{0.61} & \textbf{0.35}\\
\hline
\end{tabular}
}
\caption{\textbf{Detailed results on ETH3D SLAM~\cite{eth3d_slam}} (32 sequences not included in Table~5 of the main paper). RMSE ATE, Vertical FoV and Scale errors for three dense unconstrained (U) methods.} 
\label{tab:eth3d_full}
\end{center}
\vspace*{-1mm}
\end{table*}

\myparagraph{Number of views in memory}
The plots in Fig.~\ref{fig:ablation} suggest that the model can successfully leverage up to $n\seq 50$ views in the memory, despite being only trained with at most $10$ views. Beyond $50$ views, the results do not improve further. 
The results in Tab.~\ref{tab:spannercompar_suppmat} reinforce this observation. \munster{}-224 $n\seq 10, s\seq 1$ is worse than $n\seq all, s\seq 1$.
We note that in the \spanner{} evaluation protocol, there are only $10$ images used for DTU sequences, from $10$ to $42$ images for NRGBD and from $25$ to $50$ for 7-Scenes explaining the limited difference between $n\seq 20, s\seq 1$ and $n\seq all, s\seq 1$.

\myparagraph{Number of views at once}
We also study how \munster{} behaves when using multiple views to update the memory at the same time. Note that in most experiments we only use one image at a time, except for initialization where we start with $s{=}2$ images. We run experiments with $s\seq 5$ and $s\seq 10$ views to update the memory. Results on DTU, NRGBD and 7-Scenes are shown in Tab.~\ref{tab:spannercompar_suppmat}  for both \munster{}-224 and \munster{}-512. 
We observe that predicting multiple images at the same time does not significantly degrade the performance. Interestingly, it   
yields a decent increase in FPS for \munster{}-224: (72.59 for $s{\seq} 10$ \vs  40.41 for $s{\seq} 1$), at the cost of a slightly increased memory usage (4.8G for $s{\seq} 10$ \vs 4.1G for $s{\seq} 1$). 
For \munster{}-512, the FPS difference is not significant and does not really justify the increased memory cost.

\section{Qualitative examples}
\label{supsec:quali}
As promised in Sections \ref{sec:intro} and \ref{ssec:vo}, we provide here additional qualitative results on various datasets. Image poses are predicted via Procrustes analysis, and for visualization purposes, we only show highly confident points.  
We wish to demonstrate the ease of use and plug-and-play nature of \munster{} in varied scenarios, and its robustness indoors Fig.~\ref{fig:qualitative1_supmat}, outdoors Fig.~\ref{fig:qualitative3_supmat} or object-centric Fig.~\ref{fig:qualitative2_supmat}.

We emphasize that our approach does not assume a single set of intrinsics for an image collection and seamlessly works  with heterogeneous sensors, as shown in the ScanNet++ scenes in Fig.~\ref{fig:qualitative1_supmat} (\textit{left}).

\section{Memory Management}
\label{supsec:code}

We provide in Fig.~\ref{code:online} the pseudo-code for the online algorithm described in Sec.~\ref{sec:memory}.
We believe \munster{} to be an important step towards a simplified VO and even RGB-SLAM pipeline. In fact, the network is internally managing all the traditionally heavily engineered steps such as keypoint selection and matching, pose estimation and 3D triangulation. Our results show that a simplistic memory frame selection mechanism is enough to outperform traditional VO methods.

\begin{figure}[ttt]
\begin{lstlisting}[
    language=Python, 
    basicstyle=\footnotesize\ttfamily, 
    keywordstyle=\color{blue}, 
    commentstyle=\color{green!50!black}, 
    frame=single,          % This creates a full box around the code
    breaklines=true, 
    breakatwhitespace=true, 
    columns=fullflexible, 
    linewidth=\linewidth,
    rulecolor=\color{gray!90}
]
perc=85
thresh=0.05
def online_update(frame,munst3r,memory,scene3d):
  # forward view with current memory
  Xi1, Xii, new_tokens=munst3r(frame,memory)
  # depth and focal from local prediction
  depth = Xii[...,-1] # [H,W]
  focal = get_focal(depth)
  # pose via procrustes between local and global 
  pose = procrustes_align(Xii,Xi1) 
  # viewing direction for each pixel
  rays = normalize(Xi1-pose.t) 
  # viewing-direction aware 3D discovery rate
  dists = scene3d.query(Xi1,rays)
  if percentile(dists/depth,perc) > thresh:
    # Append frame tokens to memory
    memory.append(new_tokens)
    # Add predicted points and directions
    scene3d.add(Xi1, rays)
  return Xi1,focal,pose,depth
\end{lstlisting}
\vspace{-.5cm}
\caption{Python code for Uncalibrated Visual Odometry.}
\label{code:online}
\vspace{-.5cm}
\end{figure}

\section{Full Visual Odometry results}
\label{supsec:vo}
In this section we provide the detailed tables of evaluating online \munster{} model 
on TUM RGB-D \cite{Sturm2012ASystems} and ETH3D~\cite{eth3d_slam}.

\subsection{Full TUM RGB-D dataset}
\label{ssec:vo_tum}
Tables \ref{tab:tum_ate}, \ref{tab:tum_fov}, \ref{tab:tum_scale}, shown in Sec. \ref{ssec:vo}, reported RMSE APE, vertical FoV and scale errors for 11 sequences of TUM-RGBD dataset~\cite{Sturm2012ASystems}; these sequences were selected for being most frequently evaluated by the recent state-of-the-art methods.
Here, Tab.~\ref{tab:tum_full} completes our evaluation and presents detailed results of \munster{}-224 on all for 34 remaining TUM-RGBD sequences (see the main paper and Table \ref{tab:tum_group} for the results aggregated over 5 categories of sequences). The table also includes comparison to GlORIE-SLAM \cite{ZhangX24GlORIESLAM} run in VO mode. The full dataset contains extremely hard sequences and, to the best of our knowledge, it has never been fully tested previously by other VO works. Despite its simplicity, \munster{} performs overall on par with GlORIE-VO while being substantially faster. Furthermore, \munster{} outperforms \spanner{} on all sequences.

\subsection{ETH3D dataset}
\label{ssec:vo_eth3d}
Finally, Tab.~\ref{tab:eth3d_full} presents detailed results of \munster{}-224 on 32 ETH3D sequences which complete the evaluation of the 8 sequences included in the main paper (Tab.~\ref{tab:eth3d_ate}). The table reports RMSE APE, vertical FoV and scale errors for all 32 sequences and compares \munster{} and \munster{}-C to \spanner{}, both in $224$ resolution. As the table clearly shows, both versions of \munster{} again outperform \spanner{} on the vast majority of sequences, although we note that there is still room for improvements on some sequences, usually when the scene size becomes too large. We hope these systematic results, along with code, will help foster research in this direction. 

{
    \small
    \bibliographystyle{ieeenat_fullname}
    \bibliography{dust3r,munst3r,slam}
}

\end{document}